%% file: 0_main.tex
\documentclass[a4paper,fleqn]{cas-dc}

\usepackage[authoryear,longnamesfirst]{natbib}
\usepackage{color}
\usepackage[small]{caption}
\usepackage{amsmath,amsfonts,amssymb}
\usepackage{array}
\usepackage{textcomp}
\usepackage{stfloats}
\usepackage{url}
\usepackage{verbatim}
\usepackage{graphicx}
\usepackage{booktabs}
\usepackage{multirow}
\usepackage{algorithm}
\usepackage{algorithmic,bm}
\usepackage[normalem]{ulem}
\useunder{\uline}{\ul}{}
\def\tsc#1{\csdef{#1}{\textsc{\lowercase{#1}}\xspace}}
\tsc{WGM}
\tsc{QE}
\tsc{EP}
\tsc{PMS}
\tsc{BEC}
\tsc{DE}

\newcommand{\zf}[1]{{\color{black}{#1}}}
\def\method{LEAF}
\begin{document}
\let\WriteBookmarks\relax
\def\floatpagepagefraction{1}
\def\textpagefraction{.001}
\let\printorcid\relax


\shorttitle{\method{}: Unveiling Two Sides of the Same Coin in Semi-supervised Facial Expression Recognition}

\shortauthors{Zhang et~al.}
\title [mode = title]{\method{}: Unveiling Two Sides of the Same Coin in Semi-supervised Facial Expression Recognition}
                   



%
\author[1,2]{\textcolor{black}{Fan Zhang}}
\ead{zfkarl1998@gmail.com}

\fnmark[1]



\credit{Methodology, Data Collection, Experiment, Writing}

\affiliation[1]{organization={Shenzhen Technology University},
    city={Shenzhen},
    postcode={518118}, 
    state={Guangdong},
    country={China}}
\affiliation[2]{organization={Georgia Institute of Technology},
    city={Atlanta},
    postcode={30332}, 
    state={GA},
    country={USA}}
    
\author[3]{\textcolor{black}{Zhi-Qi Cheng}}
\fnmark[1]
\credit{Methodology, Data Collection, Writing}
\affiliation[3]{organization={Carnegie Mellon University},
    city={Pittsburgh},
    postcode={15213}, 
    state={PA},
    country={USA}}
\author[4]{\textcolor{black}{Jian Zhao}}
\credit{Data Collection}
\fnmark[1]

\affiliation[4]{organization={Institute of Artificial Intelligence, China Telecom},
    city={Beijing},
    postcode={100027}, 
    country={China}}

\author[2]{\textcolor{black}{Xiaojiang Peng}}
\credit{Methodology, Review}
\ead{pengxiaojiang@sztu.edu.cn}
\cormark[1]

\author[4]{\textcolor{black}{Xuelong Li}}
\credit{Writing, Review}
\cormark[1]

\cortext[cor1]{Corresponding authors.}

\fntext[fn1]{Equal contributions.}



\input{1_abstract}

\maketitle

\input{2_introduction}

\input{3_related_work}
\input{4_method}

\input{5_experiment}

\input{6_conclusion}


\bibliographystyle{model1-num-names}

\bibliography{7_rec}


\end{document}

%% file: 1_abstract.tex
\begin{abstract}
Semi-supervised learning has emerged as a promising approach to tackle the challenge of label scarcity in facial expression recognition (FER) task. However, current state-of-the-art methods primarily \textit{focus on one side of the coin, i.e., generating high-quality pseudo-labels}, while \textit{overlooking the other side: enhancing expression-relevant representations}. In this paper, we \textit{unveil both sides of the coin} by proposing a \textit{unified} framework termed hierarchica\underline{L} d\underline{E}coupling \underline{A}nd \underline{F}using (\method{}) to \textit{coordinate} expression-relevant representations and pseudo-labels for semi-supervised FER.
\method{} introduces a hierarchical expression-aware aggregation strategy that operates at three levels: \zf{semantic}, instance, and category.~(1)~At the \zf{semantic} and instance levels, \method{} \textit{decouples} representations into expression-agnostic and expression-relevant components, and \textit{adaptively fuses} them using learnable gating weights.~(2)~At the category level, \method{} \textit{assigns} ambiguous pseudo-labels by \textit{decoupling} predictions into positive and negative parts, and employs a consistency loss to ensure agreement between two augmented views of the same image.
Extensive experiments on benchmark datasets demonstrate that by unveiling and harmonizing both sides of the coin, \method{} outperforms state-of-the-art semi-supervised FER methods, effectively leveraging both labeled and unlabeled data. Moreover, the proposed expression-aware aggregation strategy can be seamlessly integrated into existing semi-supervised frameworks, leading to significant performance gains. 
Our code is available at \url{https://github.com/zfkarl/LEAF}.
\end{abstract}

%% file: 2_introduction.tex
\section{Introduction}
\label{sec:intro}
Facial expressions are a critical component of human communication, serving as a primary means of conveying emotions. With the increasing influence of artificial intelligence-generated content, facial expression recognition (FER) has gained significant attention in recent years, finding applications in various domains such as human-machine interaction \cite{shibata1997artificial,sun2019design,erol2019toward} and the development of digital humans \cite{volonte2021effects,loveys2021effects}. However, accurately detecting and interpreting facial expressions, particularly in supervised settings, presents substantial challenges for emotion recognition systems.

\begin{figure}[t]
    \centering
    \includegraphics[width=\linewidth]{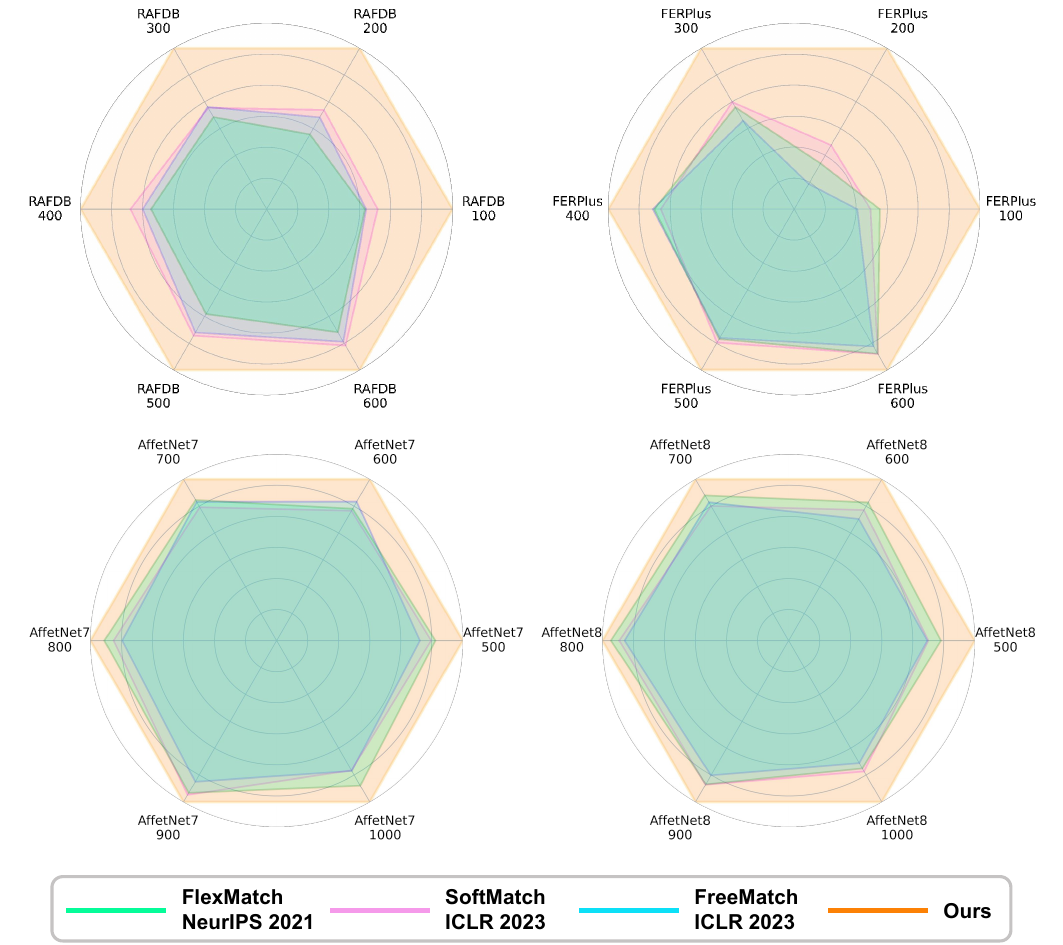}
    \caption{Our \method{} consistently outperforms state-of-the-art semi-supervised FER approaches across different settings.}
    \label{fig: ills}
\end{figure}

The main obstacle in FER stems from the difficulty in obtaining a large volume of labeled facial expression data. This challenge is compounded by high inter-class similarities and the potential for errors, even among well-trained annotators. Consequently, FER under label scarcity has emerged as a practical yet under-explored problem. While most existing FER approaches \cite{li2017reliable,li2021adaptively,she2021dive,xue2021transfer} exhibit a data-hungry nature, heavily relying on extensive labeled data, there is a pressing need for a semi-supervised FER approach~\cite{grandvalet2004semi,rosenberg2005semi,hadsell2006dimensionality,song2017semantic} that can effectively utilize a small amount of labeled data in conjunction with a large amount of unlabeled data to recognize facial expressions accurately.

Recent progress in semi-supervised FER methods \cite{adacm,margin-mix} and semi-supervised image classification approaches \cite{fixmatch,flexmatch,wang2022freematch,chen2023softmatch} has primarily focused on one side of the coin: enhancing the quality or quantity of pseudo-labels. These methods aim to generate accurate and diverse pseudo-labels for unlabeled data, which can then be used to train the model in a supervised manner. However, they often overlook the other side of the coin: the potential improvement in representation quality stemming from the inherent minor inter-class differences in FER. Learning expression-relevant representations is crucial, as the differences between various facial expressions can be subtle. The model must be capable of capturing these nuances to accurately classify emotions. By focusing solely on pseudo-labels, existing methods~\cite{song2017semantic,hadsell2006dimensionality} may fail to learn discriminative representations that can effectively distinguish between different expressions.

In this paper, we propose a new data-efficient framework called hierarchica\underline{L} d\underline{E}coupling \underline{A}nd \underline{F}using (\method{}) to address both sides of the same coin in semi-supervised FER. \method{} aims to improve the quality of representations as well as the quality and quantity of pseudo-labels from a hierarchical perspective. The core of \method{} lies in gradually teaching the network to distinguish facial expression representations and pseudo-labels into expression-agnostic and expression-relevant parts through decoupling strategies at different levels. The decoupled representations and pseudo-labels are then fused by automatically assigning different weights to them, enabling the model to focus on the parts more relevant to facial expressions, thereby achieving better recognition performance.

Specifically, \method{} introduces three levels of \textit{d\underline{E}coupling \underline{A}nd \underline{F}using (EAF) strategy}. At the \zf{semantic} and instance levels, the EAF strategy draws inspiration from the Mixture-of-Experts (MoE) technique \cite{eigen2013learning,shazeer2017outrageously,fedus2022switch}, allowing individual experts to learn and handle their respective specialized representations (i.e., expression-agnostic and expression-relevant parts). These representations are then automatically weighted and fused through a learnable gating network. At the category level, inspired by existing metric learning works \cite{liu2017sphereface,sun2020circle,wang2018additive,wang2018cosface,fpl}, an ambiguous consistency loss is designed to minimize the distance between the prediction distributions obtained in two forward processes. Unlike traditional deterministic consistency losses, \method{} assigns several candidate pseudo-labels to each unlabeled sample, ambiguously labeling them as positive (expression-relevant) or negative (expression-agnostic). A margin is used to control the distance between the positive and negative categories, enhancing the consistency between the distributions.

By simultaneously addressing both sides of the coin, \method{} aims to learn a unified and coordinated representation space that captures the subtle differences between facial expressions while generating accurate and diverse pseudo-labels. This approach not only improves the quality of representations but also ensures that the pseudo-labels are consistent with the underlying expression-relevant information. Consequently, \method{} can effectively leverage both labeled and unlabeled data to enhance the performance of semi-supervised FER. The effectiveness of \method{} is demonstrated through extensive experiments on several public and widely-used benchmark datasets, as shown in Fig. \ref{fig: ills}. Furthermore, as a plug-and-play module, the proposed EAF strategies are adaptive to other existing methods, showing improved performance when integrated.

The main contributions of this paper can be summarized as following three points:
\begin{itemize}
\item We explore a practical yet rarely investigated problem of FER under label scarcity and identify the shortcomings of existing semi-supervised approaches, which focus only on improving the quality or quantity of pseudo-labels while overlooking the enhancement of representation quality.
\item We propose a semi-supervised FER framework \method{} that automatically distinguishes between expression-agnostic and expression-relevant representations and pseudo-labels, and assigns them different weights in a hierarchical decoupling and fusing manner, effectively addressing both sides of the same coin in semi-supervised FER.
\item Extensive experiments on several benchmark datasets demonstrate that \method{} consistently outperforms a series of state-of-the-art approaches, and the proposed EAF strategies can be integrated into existing methods to boost performance.
\end{itemize}

%% file: 3_related_work.tex
\section{Related Work}
\label{sec:related}
\subsection{Facial Expression Recognition}
There have been numerous approaches proposed for FER, which can be classified into two primary research categories: methods based on handcrafted features \cite{hu2008multi,luo2013facial,pietikainen2011computer} and those based on deep learning techniques \cite{li2017reliable,li2021adaptively,she2021dive,xue2021transfer}. In traditional research, the emphasis is on extracting texture information from datasets obtained in controlled laboratory settings, such as CK+ \cite{lucey2010extended} and Oulu-CASIA \cite{zhao2011facial}. With the advent of large-scale unconstrained FER datasets \cite{barsoum2016training,li2017reliable,mollahosseini2017affectnet}, deep facial expression recognition (DFER) algorithms have emerged, aiming to create effective neural networks or loss functions that can deliver superior performance. 
For instance, Zhang et al. \cite{zhang2024leave} propose re-balanced attention maps and re-balanced smooth labels to mine extra knowledge from both major and minor classes for imbalanced FER.
Wu et al. \cite{wu2023net} leverage facial landmarks to mitigate the impact of label noise in FER.
Chen et al. \cite{chen2024cfan} propose to transfer knowledge from static images to unlabeled frames of dynamic videos.
Moreover, recent progress in FER lies in human prior-based network \cite{xie2020facial,wang2021light,li2021learning,gu2022toward,li2023fg,cai2024mfdan}, self-supervised learning techniques \cite{liu2022devil,zhang2024self}, and cross-modal prompts \cite{zhou2024ceprompt}.

However, most of the existing FER methods are data-hungry. Although several semi-supervised FER approaches \cite{adacm,margin-mix} have been proposed to explore recognizing facial expressions under label scarcity and make some progress, they only concentrate on enhancing the quality or quantity of pseudo-labels, while ignoring that the representations of facial expressions can also be improved.

\subsection{Semi-Supervised Learning}
In recent years, there has been notable progress in applying semi-supervised learning methods to tackle challenging problems \cite{fixmatch,flexmatch}. 
These methods utilize various techniques such as consistency regularization \cite{regularization,uda}, entropy minimization \cite{grandvalet2004semi,lee2013pseudo}, and traditional regularization \cite{berthelot2019mixmatch} to make effective use of unlabeled data. 
Among them, pseudo-labeling has emerged as a pioneering semi-supervised learning technique for obtaining hard labels from model predictions.
Notably, threshold-based pseudo-labeling approaches have been employed to select unlabeled samples with high-confidence predictions. 
For instance, some methods \cite{fixmatch,uda} employ a fixed threshold to obtain pseudo-labels and incorporate both weak and strong augmentations to enforce consistency regularization. 
Other methods \cite{flexmatch,xu2021dash} explore dynamic threshold strategies to adaptively determine which samples to assign pseudo-labels. 

However, almost all the threshold-based pseudo-labeling methods inevitably result in some low-confidence samples not being fully utilized.
To this end, \method{} employs EAF at the category level through an ambiguous pseudo-label selecting strategy to make full use of all the unlabeled data for consistency regularization.

\begin{figure*}[t]
    \centering
    \includegraphics[width=\linewidth]{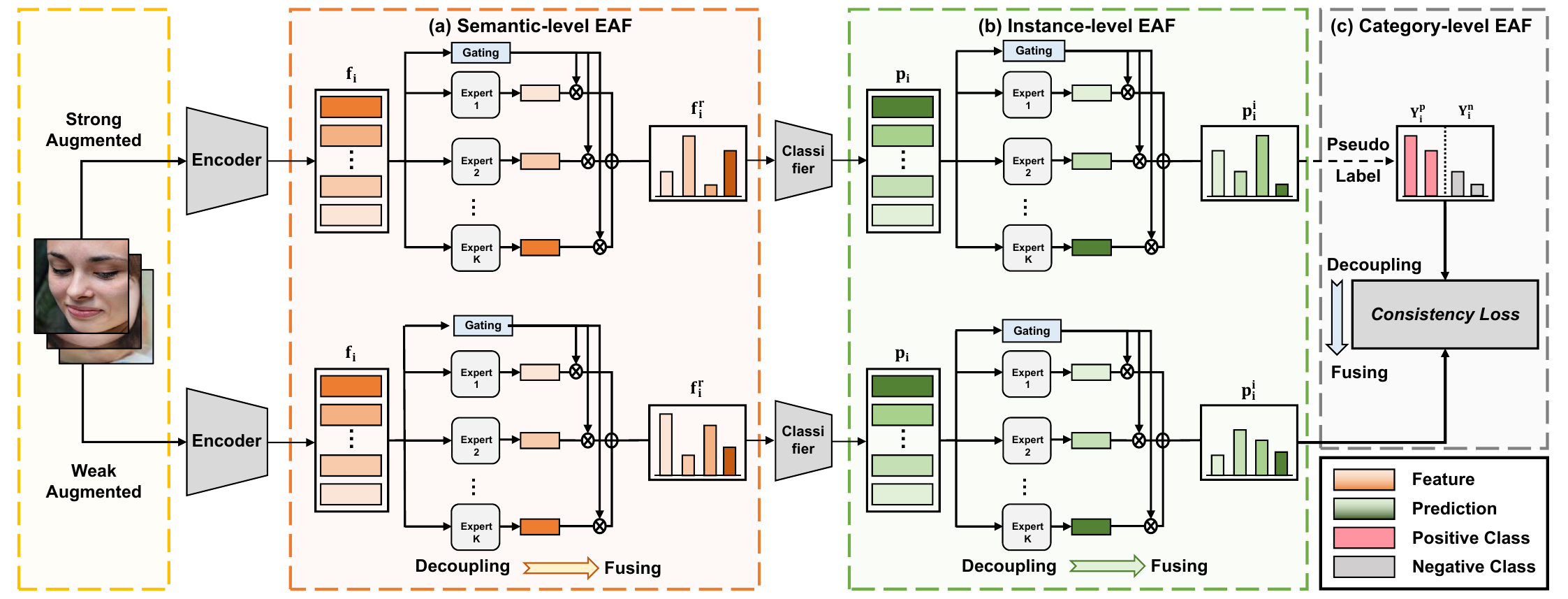}
    \caption{An overview of \method{}. The weak and strong augmented views of facial expressions are first mapped into the embedding space through a shared encoder. Then we conduct the \zf{semantic}-level EAF and the instance-level EAF before and after the classifier to reorganize weights for expression-relevant and expression-agnostic representations, respectively. After getting the predictions, we adopt the category-level EAF to generate ambiguous pseudo-labels for consistency regularization.}
    \label{framework}
\end{figure*}

%% file: 4_method.tex
\section{Problem Definition}
\label{sec:problem_def}

Facial expression recognition (FER) aims to classify human facial expressions into discrete categories. In real-world scenarios, labeled facial expression data is often scarce, while unlabeled data is abundant. Semi-supervised learning (SSL) methods leverage both labeled and unlabeled data to improve FER performance. Given an FER dataset $\mathcal{D} = \mathcal{D}^{l}\cup \mathcal{D}^{u}$, where $\mathcal{D}^{l}=\{(x_i^{l}, y_i)\}_{i=1}^{N_{l}}$ and $\mathcal{D}^{u}=\{(x_i^{u})\}_{i=1}^{N_{u}}$ denote the labeled and unlabeled samples, respectively, the goal of semi-supervised FER is to learn the parameters $\theta$ of a model $F(x; \theta)$ by optimizing a loss function that combines supervised and unsupervised terms:
\begin{equation}
\mathcal{L} = \frac{1}{N_{l}}\sum_{i=1}^{N_{l}} \mathcal{L}^{s}(F(x_i^{l}; \theta), y_i) + \frac{\lambda}{N_{u}}\sum_{i=1}^{N_{u}} \mathcal{L}^{u}(F(x_i^{u}; \theta), \Bar{y_i}),
\label{loss:overall}
\end{equation}
where $\mathcal{L}^{s}$ and $\mathcal{L}^{u}$ represent the supervised classification loss and unsupervised consistency loss, respectively, and $\lambda$ is a regularization coefficient to balance the two terms. Equation \ref{loss:overall} encapsulates the core idea of semi-supervised learning: utilizing both labeled and unlabeled data to enhance model performance. The supervised loss $\mathcal{L}^{s}$ ensures that the model learns from the annotated samples, while the unsupervised consistency loss $\mathcal{L}^{u}$ encourages consistent predictions for unlabeled samples under different augmentations. 

The state-of-the-art SSL approaches for FER \cite{adacm,fixmatch,flexmatch,chen2023softmatch,wang2022freematch} typically formulate the supervised loss as the cross-entropy loss between the model prediction $p_i$ and the label $y_i$, and the unsupervised loss as the consistency loss between the model prediction $p_i$ and the pseudo-label $\Bar{y_i}$. However, these methods suffer from two main limitations \cite{fixmatch,flexmatch,adacm,margin-mix}:
(1) \textbf{Equal treatment of representations}: Existing methods fail to consider the varying discriminative power of different facial expression representations, treating all representations equally. This approach overlooks the fact that some representations may provide more valuable information for accurate recognition than others.
(2) \textbf{Inappropriate pseudo-labeling}: The assignment of hard pseudo-labels to unlabeled samples can be problematic, especially when labeled data is scarce. The limited availability of labeled data hinders the model's ability to make accurate predictions, leading to potentially unreliable pseudo-labels.

\section{LEAF Framework}
\label{sec:framework_overview}

Our strategy, illustrated in Fig. \ref{framework}, involves implementing the EAF strategy at three distinct levels. First, augmented views of samples are encoded into deep features.~(1)~At the \zf{semantic} level, these features are distributed among various experts and fused by a gating network before reaching the classifier (Sec. \ref{sec:rep_level}).~(2)~Next, at the instance level, predictions from the classifier are decoupled by additional experts and fused through another gating network (Sec. \ref{sec:ins_level}).~(3)~Finally, at the category level, pseudo-labels are assigned to predictions, and consistency between the two forward distributions is enhanced (Sec. \ref{sec:cat_level}). Notably, the pseudo-labels are dynamic, decoupled into positive and negative labels, and fused during consistency regularization. The following sections provide detailed explanations of each component.

\subsection{\zf{Semantic}-level Decoupling and Fusing}
\label{sec:rep_level}

The \zf{semantic}-level strategy, depicted in Fig. \ref{framework} (a), assumes that encoded deep features capture rich geometric or texture information. However, we argue that this information redundancy may hinder fine-grained expression recognition, as not all details contribute equally to the performance of final predictions. Inspired by the powerful Mixture-of-Experts (MoE) technique \cite{eigen2013learning,shazeer2017outrageously,fedus2022switch}, we propose decoupling these features, allowing the network to autonomously determine the usefulness of each feature for expression recognition. As a result, the network learns to assign greater weights to more impactful features.

\begin{figure}[t]
    \centering
    \includegraphics[width=\linewidth]{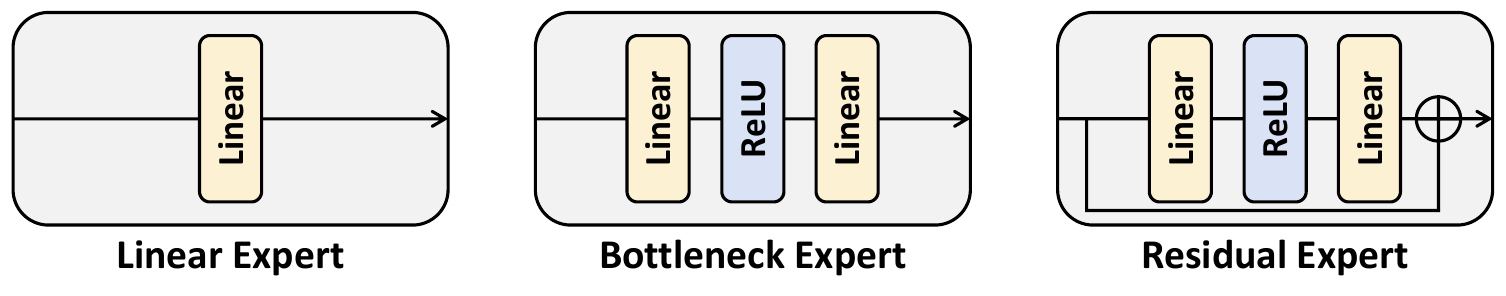}
    \caption{The detailed structure of linear expert, bottleneck expert, and residual expert.}
    \label{framework2v2}
\end{figure}

Specifically, at the \zf{semantic} level, the deep features $f_i$ are dispatched to several experts and fused by an additional gating network. The output of the \zf{semantic}-level EAF can be formulated as:
\begin{equation}\label{s_EAF}
f_{i}^{r} = \sum_{j=1}^{n} G(f_i)\cdot E_{j}(f_i),
\end{equation}
where $n$ refers to the number of experts $E_j$ involved in processing the representations of expression information. The gating network $G(\cdot)$ is a feed-forward network (FFN) with learnable weights $w^{r}$. Additionally, we design three types of experts, whose structures are shown in Fig. \ref{framework2v2}. The performance of different experts will be further discussed in Sec. \ref{sec: ablation_expert}.

The gating network plays a crucial role in this process by learning to assign appropriate weights to the outputs of each expert. The use of a trainable FFN as the gating network provides the flexibility to learn complex mappings between the input features and the expert weights, allowing our model to dynamically adapt the fusion strategy based on the characteristics of the input data, enabling more effective utilization of the available information.

To direct experts' attention towards expression-relevant features, we aim to limit the number of experts with non-zero weights, specifically for handling expression-relevant features in distinct scenarios. The weights of the remaining experts are omitted in the calculation process. We achieve this by sampling the top $K$ outputs of the experts and aggregating them through gating:
\begin{equation}
G(f_i) = \delta (TopK(\sigma( w^{r}\cdot f_i))),
\end{equation}
where $\delta$ and $\sigma$ denote softmax activation and softplus activation, respectively. For experts whose outputs are not within the $TopK$ values, their representation values are set to $-\infty$. These values will become zero after applying the softmax function.

This top-$K$ gating mechanism is a key innovation that enables our model to focus on the most relevant experts for each input sample. By dynamically selecting the top $K$ experts based on their activation values, we ensure that only the most informative features are used for the final prediction. This contrasts with traditional approaches that either use all experts equally or rely on fixed, hand-crafted rules for expert selection.

\subsection{Instance-level Decoupling and Fusing}
\label{sec:ins_level}

Building upon the \zf{semantic}-level strategy, we employ a classifier $C(\cdot)$ to project the deep features into a low-dimensional space, yielding the final predictions $p_i= C(f_i^{r})$. We argue that the high inter-class similarity characteristic of facial expressions persists not only in the high-dimensional feature space but also in the low-dimensional category distribution space (Fig. \ref{framework} (b)). Therefore, extending the EAF strategy to the instance level, as defined in Eqn. \ref{i_EAF}, is expected to enhance recognition performance.

\begin{equation}\label{i_EAF}
p_{i}^{i} = \sum_{j=1}^{n} G(p_i)\cdot E_{j}(p_i).
\end{equation}
Similarly, the gating network can be described as:
\begin{equation}
G(p_i) = \delta (TopK(\sigma( w^{i}\cdot p_i))).
\end{equation}
It is worth noting that the gating networks $G(\cdot)$ of the \zf{semantic}-level and instance-level share the same architecture but have different parameters. The weights $w^{r}$ and $w^{i}$ are distinct and can be simultaneously optimized during the learning process.

By applying the EAF strategy at the instance level, our approach enables the model to leverage the complementary information captured by different experts to improve the final predictions. The gating network learns to assign appropriate weights to the outputs of each expert, effectively combining their predictions based on their relative importance for each input instance.

The use of top-$K$ gating at the instance level serves a similar purpose as in the \zf{semantic} level, encouraging the model to focus on the most relevant experts for each input sample. This helps to mitigate the impact of noisy or ambiguous predictions that may arise due to the high inter-class similarity of facial expressions.
Another key advantage of the instance-level EAF strategy is that it allows the model to adaptively refine the predictions based on the specific characteristics of each input sample. By learning to assign different weights to the experts for different instances, our approach can effectively capture the subtle variations in facial expressions that may be crucial for accurate recognition.

\subsection{Category-level Decoupling and Fusing}\label{sec:cat_level}

Several approaches \cite{fixmatch,uda,flexmatch,xu2021dash} assign hard pseudo-labels to unlabeled data, aiming to convert unsupervised learning scenarios into supervised ones. However, most of these methods employ a threshold to filter out low-confidence pseudo-labels, utilizing only a portion of unlabeled samples while discarding the rest. Moreover, considering the subtle inter-class differences in FER, fixed pseudo-labels may encounter challenges such as incorrect assignments and discarding useful information from highly similar negative classes.

To address these challenges, we introduce the EAF strategy at the category level, as shown in Fig. \ref{framework} (c). This involves decoupling the probability distribution of expression categories into positive and negative classes and subsequently fusing them for consistency regularization. First, we employ the softmax function to convert the prediction of a strong augmented sample $p_{i}^{i}$ into a soft label distribution $y_{i}$. Assuming there are $k$ classes in total, we decouple these label distributions and maintain a positive set $Y_{sa}^{p}$ and a negative set $Y_{sa}^{n}$ for each strong augmented sample:
\begin{align}
Y_{sa}^{p} &= \{y_{sa}^{1}, y_{sa}^{2},..., y_{sa}^{m}\}, \\
Y_{sa}^{n} &= \{y_{sa}^{m+1}, y_{sa}^{m+2},..., y_{sa}^{k}\},
\end{align}
where $m$ denotes the number of positive emotion classes, and the remaining ones are considered as negative classes. Note that the positive and negative emotion classes here are not pre-defined knowledge in affective computing. Instead, we use the cumulative probability of the sorted prediction to determine the value of $m$:
\begin{equation}
\sum_{j=1}^{m} \psi(y_{sa}^{j}) \ge T,
\end{equation}
where $\psi$ denotes the sort process, and $T$ is a threshold set to $0.9$ empirically. Inspired by metric learning works \cite{liu2017sphereface,sun2020circle,wang2018additive,wang2018cosface,fpl}, we force the prediction of weak augmented sample to be consistent with the above partition of the strong augmented sample with a learning objective as:
\begin{equation} \label{eq:minmax}
min (Y_{wa}^{p}) - max (Y_{wa}^{n}) > \epsilon \geq 0,
\end{equation}
where $\epsilon \geq 0$ is a margin to control the distance between the two sets, and we set $\epsilon = 0$ by default. Eqn. \ref{eq:minmax} ensures all the candidate labels in the positive set have higher scores than the others in the negative set. Next, we fuse these label distributions by taking the negative form of Eqn. \ref{eq:minmax} as the loss function:
\begin{equation}
\mathcal{L}^{U} = \eta ( max (Y_{wa}^{n}) - min (Y_{wa}^{p}) ),
\end{equation}
where $\eta$ denotes the $ReLU$ activation.

However, the $max$ and $min$ functions are non-differentiable, so we approximate them into differentiable formats:
\begin{align}
&min (Y_{wa}^{p}) \approx -log(\sum_{i=1}^{m} e^{-y^{i}}), \\
&max (Y_{wa}^{n}) \approx log(\sum_{j=m+1}^{k} e^{y^{j}}), \\
&\eta (Y) = max(Y,0) \approx log(1+e^{Y}).
\end{align}
Based on these approximations, the overall consistency regularization loss is converted to:
\begin{equation} \label{d_EAF}
\begin{aligned}
\mathcal{L}^{U} &= max(max (Y_{wa}^{n}) - min (Y_{wa}^{p}),0)  \\
& = log(1+ e^{max (Y_{wa}^{n}) - min (Y_{wa}^{p})})\\
& = log(1+ e^{log(\sum_{i=1}^{m} e^{-y^{i}}) +log(\sum_{j=m+1}^{k} e^{y^{j}})} \\
& = log(1+ \sum_{i\in Y_{wa}^{p}} e^{-y^{i}} \times \sum_{j\in Y_{wa}^{n}} e^{y^{j}} ).
\end{aligned}
\end{equation}
The step-by-step training algorithm of \method{} is listed in Alg. \ref{alg1}. 

{\footnotesize
\begin{algorithm}[t]
\caption{Training Algorithm of LEAF}
\label{alg1}
\begin{algorithmic}[1]
\REQUIRE Labeled dataset $\mathcal{D}^{l}$, unlabeled dataset $\mathcal{D}^{u}$, number of training epochs $E$
\ENSURE Learned model parameters $\theta$
\STATE Initialize the model parameters $\theta$
\FOR{$e = 1$ to $E$}
    \REPEAT
        \STATE Sample mini-batch of labeled data $\{(x_i^l, y_i)\}$ from $\mathcal{D}^{l}$
        \STATE Sample mini-batch of unlabeled data $\{x_i^u\}$ from $\mathcal{D}^{u}$
        \FOR{each labeled sample $(x_i^l, y_i)$}
            \STATE Compute \zf{semantic}-level features $f_i^l$ using the model $F(\theta)$
            \STATE Apply \zf{semantic}-level EAF to obtain fused features ${f_i^l}^r$ (Eqn. \ref{s_EAF})
            \STATE Compute instance-level predictions $p_i^l$ using the classifier $C$
            \STATE Apply instance-level EAF to obtain fused predictions ${p_i^l}^i$ (Eqn. \ref{i_EAF})
            \STATE Compute supervised loss $\mathcal{L}_i^s$ using ${p_i^l}^i$ and $y_i$
        \ENDFOR
        \FOR{each unlabeled sample $x_i^u$}
            \STATE Compute \zf{semantic}-level features $f_i^u$ using the model $F(\theta)$
            \STATE Apply \zf{semantic}-level EAF to obtain fused features ${f_i^u}^r$ (Eqn. \ref{s_EAF})
            \STATE Compute instance-level predictions $p_i^u$ using the classifier $C$
            \STATE Apply instance-level EAF to obtain fused predictions ${p_i^u}^i$ (Eqn. \ref{i_EAF})
            \STATE Apply category-level EAF to obtain positive and negative sets $Y_i^p$, $Y_i^n$
            \STATE Compute consistency loss $\mathcal{L}_i^u$ using $Y_i^p$, $Y_i^n$ (Eqn. \ref{d_EAF})
        \ENDFOR
        \STATE Compute total loss $\mathcal{L} = \frac{1}{N_l}\sum_i \mathcal{L}_i^s + \frac{\lambda}{N_u}\sum_i \mathcal{L}_i^u$
        \STATE Update model parameters $\theta$ by minimizing $\mathcal{L}$
    \UNTIL{end of epoch}
\ENDFOR
\end{algorithmic}
\end{algorithm}
}

%% file: 5_experiment.tex
\section{Experiments}
\label{experiments}

\begin{table*}[t]
\centering
\caption{Comparison with state-of-the-art methods with varying numbers of labels on four benchmark datasets. The best results are shown in \textbf{boldface} and the second best results are {\ul underlined}.}
\label{exp: main}
\resizebox{\linewidth}{!}{
\begin{tabular}{l|ccc|ccc|ccc|ccc}
\toprule
Dataset & \multicolumn{3}{c|}{RAFDB} & \multicolumn{3}{c|}{FERPlus} & \multicolumn{3}{c|}{AffectNet7} & \multicolumn{3}{c}{AffectNet8} \\ \midrule
Label & 100  & 200  & 400  & 100  & 200  & 400  & 500  & 1000  & 2000  & 500  & 1000  & 2000  \\ \midrule
Pi Model \cite{laine2016temporal} & 43.39 & 52.92 & 60.29 & 42.62 & 48.18 & 58.06 & 44.76 & 50.34 & 53.26 & 39.71 & 44.26 & 46.88 \\
Pseudo-Label \cite{lee2013pseudo}& 44.36 & 52.04 & 59.61 & 39.75 & 48.03 & 55.19 & 44.85 & 48.28 & 52.28 & 40.70 & 43.03 & 47.49 \\
VAT \cite{miyato2018virtual}& 33.28 & 48.21 & 57.14 & 40.07 & 48.40 & 51.95 & 43.46 & 47.68 & 51.79 & 38.02 & 43.13 & 47.54 \\
UDA \cite{uda}& 49.22 & 59.96 & 65.80 & 44.36 & 52.48 & 62.80 & 47.69 & 51.49 & 55.48 & 42.74 & 46.38 & 48.73 \\
MeanTeacher \cite{tarvainen2017mean}& 39.50 & 52.26 & 62.71 & 37.86 & 47.43 & 58.40 & 46.98 & 50.47 & 54.09 & 41.76 & 45.87 & 49.57 \\
MixMatch \cite{berthelot2019mixmatch}& 40.40 & 54.77 & 62.62 & 43.72 & 51.06 & 56.64 & 45.20 & 49.12 & 53.82 & 41.12 & 44.07 & 49.05 \\
ReMixMatch \cite{berthelot2019remixmatch}& 39.52 & 54.22 & 60.93 & 40.63 & 48.30 & 58.15 & 43.63 & 50.80 & 54.02 & 42.66 & 45.34 & 50.17 \\
FixMatch \cite{fixmatch}& 48.13 & 60.23 & 65.04 & 48.05 & 52.77 & 60.38 & 47.83 & 51.39 & 54.75 & 42.69 & 46.04 & 49.50 \\
DeFixMatch \cite{schmutz2022don}& 50.37 & 59.21 & 65.81 & 45.08 & 53.01 & 62.14 & 47.51 & 51.32 & 55.00 & 43.28 & 46.49 & 48.49 \\
Dash \cite{xu2021dash}& 49.95 & 59.27 & 66.44 & 43.10 & 55.56 & 62.74 & 48.03 & 51.65 & 54.49 & 42.33 & 45.69 & 50.06 \\
CoMatch \cite{li2021comatch}& 49.09 & 60.81 & 65.18 & 41.24 & 49.32 & 59.74 & 47.97 & 51.88 & 55.07 & 43.94 & 46.47 & 49.03 \\
SimMatch \cite{zheng2022simmatch}& 50.47 & 58.56 & 66.83 & 45.93 & 49.07 & 60.61 & {\ul 48.06} & 52.04 & 56.09 & 43.71 & 47.40 & 50.38 \\
AdaMatch \cite{berthelot2021adamatch}& 48.35 & 57.10 & 63.61 & 43.87 & 50.48 & 59.92 & 47.03 & 51.13 & 54.23 & 41.69 & 45.43 & 49.12 \\
FlexMatch \cite{flexmatch}& 48.79 & 53.24 & 61.16 & 43.75 & 47.90 & 58.23 & 48.00 & 52.23 & 54.76 & 42.98 & 46.45 & 49.02 \\
FreeMatch \cite{wang2022freematch}& 48.89 & 55.26 & 62.09 & 41.85 & 45.57 & 58.39 & 46.74 & 50.77 & 54.46 & 41.98 & 45.96 & 49.44 \\
SoftMatch \cite{chen2023softmatch}& 49.95 & 56.13 & 63.45 & 42.97 & 49.97 & 57.58 & 47.65 & 50.71 & 54.56 & 42.06 & 46.73 & 49.26 \\
AdaCM \cite{adacm}& {\ul 56.15} & {\ul 62.82} & {\ul 67.52} & {\ul 52.11} & {\ul 58.94} & 60.12 & 47.78 &  52.94 & 56.00 & {\ul 44.29} & 46.05 & 51.24 \\ 
LION \cite{du2023lion}& 56.08 & 61.81 & 67.49 & 49.87 & 58.77 & \textbf{65.27} & 47.96 & {\ul53.13} & {\ul56.67} & 43.99 & {\ul48.07} & {\ul51.48} \\
\midrule
\method{} & \textbf{ 56.83} & \textbf{ 63.43} & \textbf{ 69.00} & \textbf{ 52.20} & \textbf{ 61.00} & {\ul 62.92} & \textbf{ 50.21} & \textbf{ 53.87} & \textbf{ 56.84} & \textbf{ 45.37} & \textbf{ 49.53} & \textbf{ 52.34} \\ \bottomrule
\end{tabular}
}
\end{table*}
\begin{figure*}[t]
    \centering
    \includegraphics[width=\linewidth]{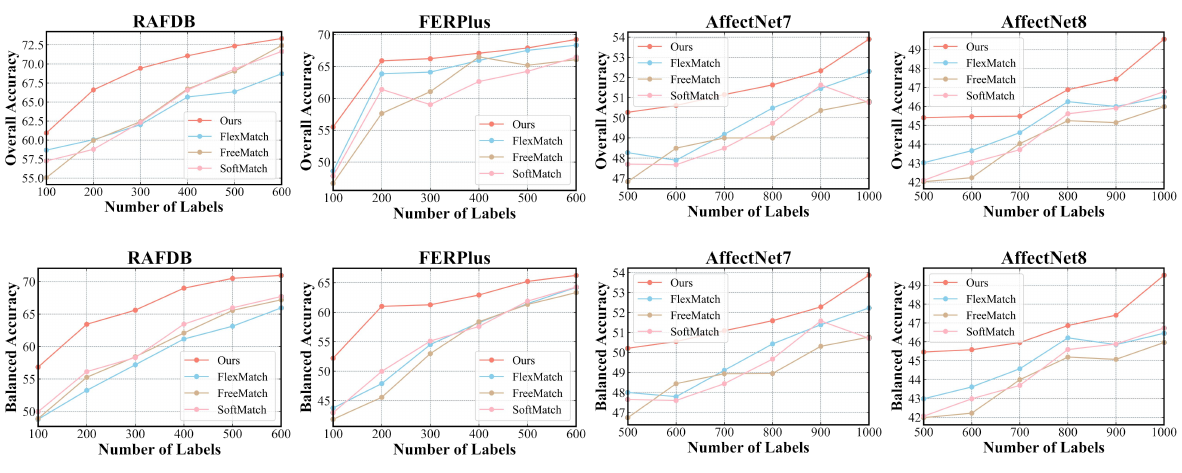}
    \caption{Performance comparison about overall accuracy and balanced accuracy with respect to different numbers of labels.}
    \label{fig: num_labels}
\end{figure*}

\subsection{Datasets and Evaluation Metrics}
We conduct extensive experiments on several widely used benchmark datasets that vary in granularity and size.
\textbf{RAFDB} \cite{li2017reliable} consists of facial images from seven distinct expressions. It is divided into a training set containing 12,271 images and a testing set with 3,068 images.
\textbf{FERPlus} \cite{barsoum2016training} comprises eight facial expressions, with a total of 24,941 training images, 3,589 validation images, and 3,589 test images.
\textbf{AffectNet} \cite{mollahosseini2017affectnet} is a large-scale dataset containing 286,564 training images and 4,000 test images, all manually annotated with eight expression labels. We consider two versions of AffectNet: \textbf{AffectNet7} and \textbf{AffectNet8}. AffectNet7 excludes the expression category of contempt, which consists of 3,667 training images and 500 test images.

Considering the imbalanced nature of the datasets, we argue that balanced accuracy is a more appropriate evaluation metric for the FER task than overall accuracy. Balanced accuracy takes into account the successful recognition of both major and minor classes \cite{zhang2024leave}, providing a more comprehensive assessment of the model's performance. Unless otherwise specified, all reported results in this paper are based on balanced accuracy.

\subsection{Implementation Details}
We implement our proposed \method{} using the PyTorch framework and conduct all experiments on NVIDIA Tesla A100 GPUs. The facial images are aligned and resized to $224\times224$ pixels. We employ the Adam optimizer with a learning rate of $5e-4$, a batch size of 64, and train the models for 20 epochs. For a fair comparison, we select ResNet-18 \cite{he2016deep} pre-trained on MS-Celeb-1M \cite{guo2016ms} as the backbone network for all baselines and our proposed method. The hyperparameters of each baseline are set according to their respective papers, and we extend these methods to the FER domain. It is important to note that we do not compare our \method{} with fully supervised FER methods, as these approaches require large amounts of labeled data and are unable to utilize unlabeled facial expression samples to improve recognition performance.

\subsection{Quantitative Comparison}

Tab. \ref{exp: main} presents a comprehensive quantitative comparison of our proposed \method{} with state-of-the-art semi-supervised learning methods on four widely used FER benchmark datasets. From these results, we can draw several key observations.
\textbf{Firstly}, it is evident that semi-supervised FER methods, such as AdaCM \cite{adacm}, LION \cite{du2023lion}, and our \method{}, outperform semi-supervised image classification methods in most cases. This can be attributed to the fact that these FER-specific approaches address the unique challenges of subtle inter-class differences among facial expressions. By effectively utilizing unlabeled data through techniques like contrastive learning for low-confidence samples and hierarchical integration of EAF modules, these methods can extract more discriminative representations and improve overall performance.
\textbf{Secondly}, we observe that the recognition of seven emotions generally yields better results compared to the recognition of eight emotions in the semi-supervised setting. This can be explained by the subtle distinction between the newly added emotion of contempt and the existing basic emotions of anger and disgust. Distinguishing between these emotions when limited labeled data is available poses a significant challenge, as the nuances can be difficult to capture without sufficient training examples.
\textbf{Finally}, and most importantly, our proposed \method{} consistently outperforms all other approaches across various settings on the four benchmark datasets. This impressive performance can be attributed to several key components of our method. The \zf{semantic}-level and instance-level EAF modules play a crucial role in extracting expression-relevant representations, while the category-level EAF strategy enables the assignment of proper pseudo-labels to ambiguous samples. The synergistic combination of these modules allows \method{} to effectively leverage both labeled and unlabeled data, resulting in robust and superior performance. A more detailed analysis of each module is provided in Sec. \ref{sec: component}.

\begin{table*}[t]
\centering
\caption{Ablation on the structure of experts across various scenarios. The best results are shown in \textbf{boldface}.}
\label{experts type}
\resizebox{0.9\linewidth}{!}{
\begin{tabular}{l|ccc|ccc|ccc|ccc}
\toprule
Dataset & \multicolumn{3}{c|}{RAFDB} & \multicolumn{3}{c|}{FERPlus} & \multicolumn{3}{c|}{AffectNet7} & \multicolumn{3}{c}{AffectNet8} \\ \midrule
Label & 100 & 200 & 400 & 100 & 200 & 400 & 500 & 1000 & 2000 & 500 & 1000 & 2000 \\
\midrule
Linear Expert & 56.01 & 60.35 & 67.18 & 51.79 & 58.72 & 62.25 & 49.19 & 52.99 & 56.71 & 44.62 & 47.84 & 52.03 \\
Bottleneck Expert & 56.23 & 62.91 & 67.44 & 51.81 & 57.22 & 62.85 & 48.89 & 53.86 & 56.80 & 45.26 & 48.10 & 52.27 \\
Residual Expert & \textbf{56.83} & \textbf{63.43} & \textbf{69.00} & \textbf{52.20} & \textbf{61.00} & \textbf{62.92} & \textbf{50.21} & \textbf{53.87} & \textbf{56.84} & \textbf{45.37} & \textbf{49.53} & \textbf{52.34}\\ \bottomrule
\end{tabular}
}
\end{table*}
\begin{table*}[t]
\centering
\caption{Ablation on proposed components on AffectNet7 with different settings. The best results are shown in \textbf{boldface}.}
\label{tab: ablation}
\resizebox{0.9\linewidth}{!}{%
\begin{tabular}{l|ccc|cc|cc|cc}
\toprule
\multirow{2.5}{*}{Model Variants} & \multicolumn{3}{c|}{LEAF} & \multicolumn{2}{c|}{Consistency Loss} & \multicolumn{2}{c|}{Overall Accuracy} & \multicolumn{2}{c}{Balanced Accuracy} \\
\cmidrule{2-10}
 & S-level & I-level & C-level & Ambiguous & CE & 500 Labels & 1000 Labels & 500 Labels & 1000 Labels \\
\midrule
\method{} w/o \zf{S-EAF} &  & \checkmark & \checkmark & \checkmark &  & 49.76 & 53.53 & 49.71 & 53.51 \\
\method{} w/o I-EAF & \checkmark &  & \checkmark & \checkmark &  & 48.82 & 52.36 & 48.76 & 52.33 \\
\method{} w/o C-EAF & \checkmark & \checkmark &  &  &  & 46.53 & 50.94 & 46.47 & 50.90 \\
\method{} w $\mathcal{L}^{CE}$ & \checkmark & \checkmark & \checkmark &  & \checkmark & 49.27 & 52.60 & 49.21 & 52.58 \\
\method{} (Full Model) & \checkmark & \checkmark & \checkmark & \checkmark &  & \textbf{50.27} & \textbf{53.90} & \textbf{50.21} & \textbf{53.87} \\ \bottomrule
\end{tabular}%
}
\end{table*}

\subsection{Qualitative Analysis}
We conduct a qualitative analysis to gain further insights under varying degrees of label scarcity. Fig. \ref{fig: num_labels} presents a visual comparison of \method{} and three state-of-the-art methods, showcasing their performance as the number of labeled samples increases.
On the RAFDB and FERPlus datasets, we gradually increase the label quantity from 100 to 500. Across this range, \method{} consistently outperforms the other methods, demonstrating its ability to effectively leverage limited labeled data. The performance gap between \method{} and the competing methods remains significant even as the number of labeled samples increases, highlighting the robustness of our approach. Given the larger scale of the AffectNet dataset, we evaluate the methods with label quantities ranging from 500 to 1000 on both AffectNet7 and AffectNet8. The results on these datasets are consistent with our observations on RAFDB and FERPlus, with \method{} maintaining its superior performance across the entire range of label quantities. Notably, the balanced accuracy of \method{} exhibits stable improvements over the other methods, indicating its effectiveness in handling class imbalance.
The qualitative analysis in Fig. \ref{fig: num_labels} provides a clear visual representation of the superiority of \method{} compared to state-of-the-art methods. As the number of labeled samples increases, all methods show improved performance, which is expected given the increased availability of annotated data. However, \method{} consistently maintains a clear advantage over the competing methods across all datasets and label quantities. This can be attributed to our approach's ability to effectively utilize unlabeled samples through the proposed EAF modules, which enable the extraction of expression-relevant representations and the assignment of accurate pseudo-labels to ambiguous samples.

\begin{figure*}[t]
    \centering
    \includegraphics[width=\linewidth]{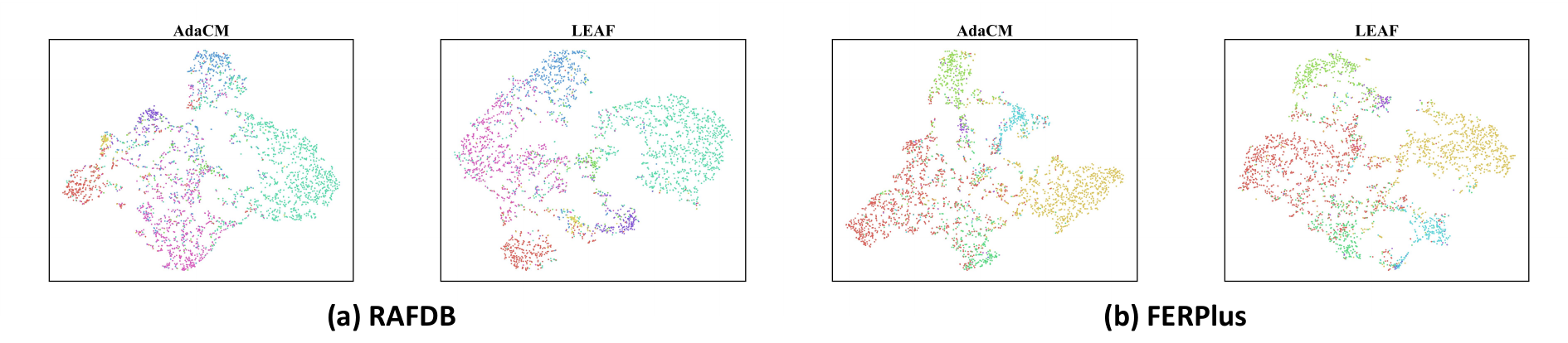}
    \caption{The t-SNE visualization with 1600 labeled samples on RAFDB and FERPlus.}
    \label{fig: tsne}
\end{figure*}
\begin{figure}[t]
    \centering
    \includegraphics[width=\linewidth]{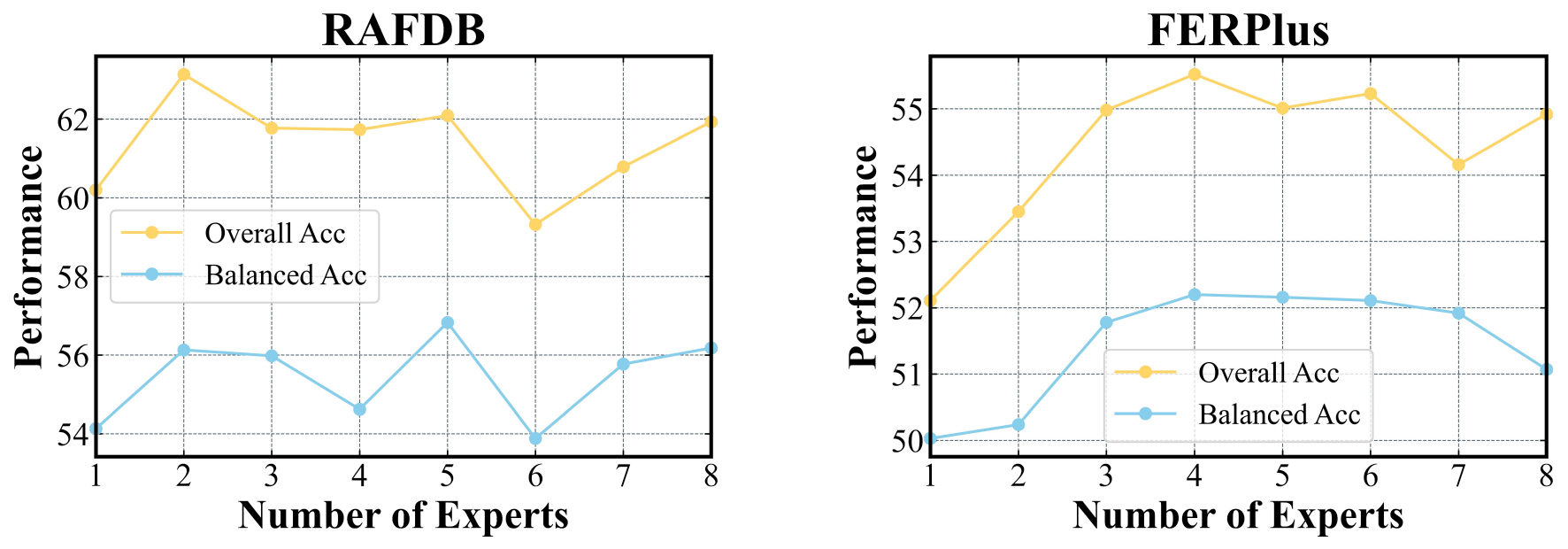}
    \caption{Sensitivity analysis of the number of experts with 100 labeled samples on RAFDB and FERPlus.}
    \label{fig: num_experts}
\end{figure}

\subsection{Ablation Studies and Analysis}

\subsubsection{Impact of Expert Structure}\label{sec: ablation_expert}

To investigate the impact of the expert structure on the performance of \method{}, we design and evaluate three different types of experts: linear expert, bottleneck expert, and residual expert (Fig. \ref{framework2v2}). The results are presented in Tab. \ref{experts type}.
We observe that the bottleneck expert, which introduces non-linear structures, enables the model to capture more complex facial expression representations compared to the linear expert, resulting in improved performance. Furthermore, the residual expert, which incorporates residual connections on top of the bottleneck structure, effectively mitigates the potential overfitting of similar facial expressions, leading to further performance gains. These findings are consistent across different settings on all four datasets, validating the effectiveness of the residual expert structure in \method{}.

\subsubsection{Contribution of Proposed Components}\label{sec: component}

To assess the individual contributions of the proposed components in \method{}, we conduct an ablation study by evaluating several model variants, as shown in Tab. \ref{tab: ablation}. \method{} w/o \zf{S-EAF}, \method{} w/o I-EAF, and \method{} w/o C-EAF denote the variants where we remove the EAF strategy at the \zf{semantic} level, instance level, and category level, respectively. Additionally, \method{} w $\mathcal{L}^{CE}$ represents the variant where we replace the proposed consistency loss with the standard cross-entropy loss.
The results demonstrate that removing the EAF strategies at any level leads to a performance decline, with the category-level EAF having the most significant impact. This highlights the importance of utilizing consistency regularization with ambiguous pseudo-labels for handling ambiguous facial expressions. Moreover, by comparing the third and fourth rows of Tab. \ref{tab: ablation}, we observe that the ambiguous consistency loss is more suitable for the semi-supervised FER task compared to the standard cross-entropy loss. Finally, the full model, which integrates all the proposed components, achieves the best performance across all settings.

\subsubsection{Sensitivity to the Number of Experts}

We analyze the sensitivity of \method{} to the number of experts, as shown in Fig.~\ref{fig: num_experts}. When the number of experts is set to $1$, the performance of EAF degenerates to that of a single projection network. As the number of experts increases, we observe performance improvements in most cases compared to the single expert scenario, with the exception of the specific case when the number of experts is $6$ on RAFDB. The optimal results are achieved with $2$ experts on RAFDB and $4$ experts on FERPlus, suggesting that the choice of the number of experts may depend on the characteristics of the dataset.

\begin{figure}[t]
    \centering
    \includegraphics[width=\linewidth]{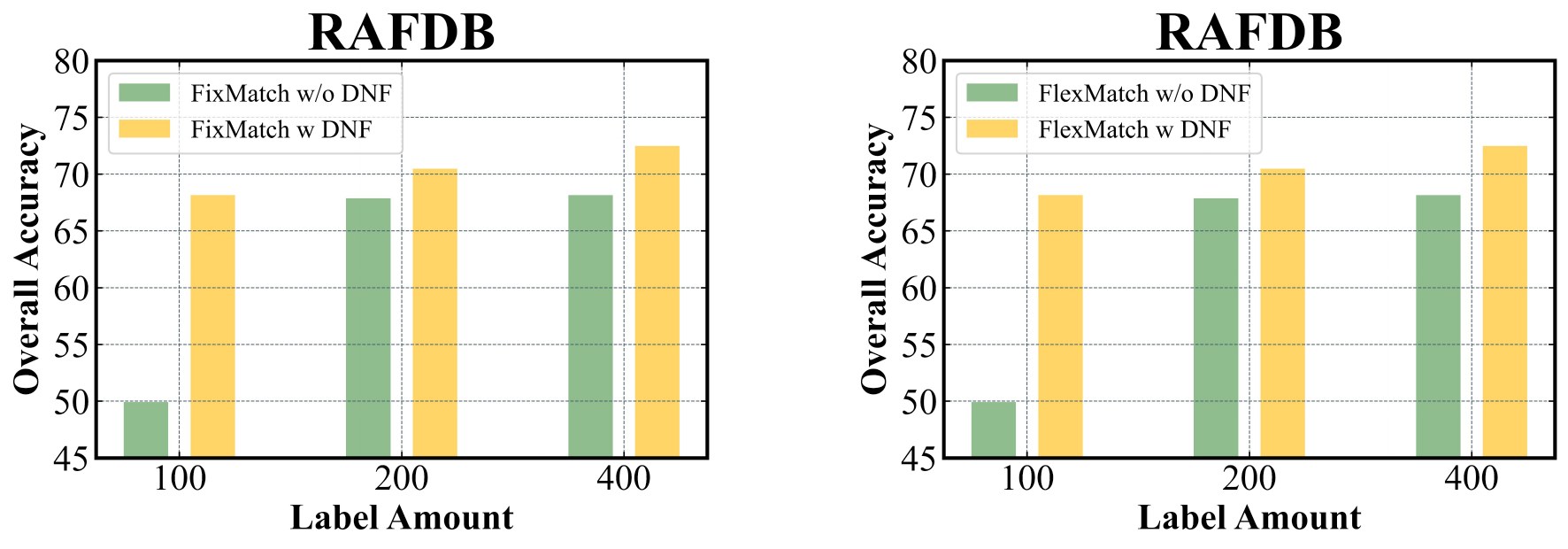}
    \caption{Performance before and after integrating our EAF strategies into other methods on RAFDB.}
    \label{fig: fusion}
\end{figure}


\subsubsection{Integration with Other Semi-Supervised Methods}

To demonstrate the flexibility and effectiveness of our proposed EAF strategies, we integrate them with two state-of-the-art semi-supervised methods, FixMatch \cite{fixmatch} and FlexMatch \cite{flexmatch}. The results, presented in Fig. \ref{fig: fusion}, show that our EAF strategies consistently bring improvements when the label quantity varies from $100$ to $300$. Notably, when the number of labels is extremely small (i.e., 100 labels), the EAF strategies provide even greater improvements, highlighting their effectiveness in scenarios with severely limited labeled data.


\subsection{Visualization Analysis}
To gain further insights into the learned feature representations, we perform t-SNE \cite{tsne} visualization of the features extracted from the encoder. Fig.~\ref{fig: tsne} presents the visualization results, where each facial expression is represented with a different color.
The visualization results demonstrate that even under label scarcity, both AdaCM \cite{adacm} and our \method{} are able to maintain good discriminability between different expressions. This indicates that these methods are capable of learning meaningful and separable feature representations for facial expressions, despite the limited availability of labeled data.
It is important to note that AdaCM \cite{adacm} achieves this discriminability by employing a contrastive learning objective, which aims to increase the distances between different expressions in the feature space. In contrast, our \method{} utilizes EAF strategies to focus the network's attention on expression-related information. These two distinct representation learning strategies demonstrate the effectiveness of different approaches in tackling the FER task.
Upon closer inspection of the visualization results, we observe that our approach slightly outperforms AdaCM \cite{adacm} in terms of the compactness and separability of the learned feature representations. This suggests that the hierarchical decoupling and fusing strategy employed by \method{}, along with the ambiguous pseudo-label generation strategy, enables the model to capture more discriminative and expression-relevant representations.

%% file: 6_conclusion.tex
\section{Conclusion}
\label{sec:conclusion}
This paper investigates a rarely explored yet practical problem of FER under label scarcity and proposes a novel semi-supervised FER approach dubbed \method{}. By decoupling and fusing the features and predictions of expressions, \method{} enables the model to focus more on expression-relevant representations automatically. Moreover, \method{} introduces an ambiguous pseudo-label generation strategy to assign expression-relevant pseudo-labels to samples, further enhancing the model's ability to learn from unlabeled data.
Extensive experiments on benchmark datasets demonstrate the effectiveness and robustness of our approach. The quantitative and qualitative results consistently show that \method{} outperforms state-of-the-art methods under various label scarcity scenarios, highlighting its potential for real-world applications.
Looking ahead, we plan to extend our approach to more challenging scenarios, such as dynamic video-based FER and multimodal emotion recognition. 


%% file: 0_main.bbl
\begin{thebibliography}{63}
\expandafter\ifx\csname natexlab\endcsname\relax\def\natexlab#1{#1}\fi
\providecommand{\url}[1]{\texttt{#1}}
\providecommand{\href}[2]{#2}
\providecommand{\path}[1]{#1}
\providecommand{\DOIprefix}{doi:}
\providecommand{\ArXivprefix}{arXiv:}
\providecommand{\URLprefix}{URL: }
\providecommand{\Pubmedprefix}{pmid:}
\providecommand{\doi}[1]{\href{http://dx.doi.org/#1}{\path{#1}}}
\providecommand{\Pubmed}[1]{\href{pmid:#1}{\path{#1}}}
\providecommand{\bibinfo}[2]{#2}
\ifx\xfnm\relax \def\xfnm[#1]{\unskip,\space#1}\fi
\bibitem[{Shibata et~al.(1997)Shibata, Yoshida, and Yamato}]{shibata1997artificial}
\bibinfo{author}{T.~Shibata}, \bibinfo{author}{M.~Yoshida}, \bibinfo{author}{J.~Yamato},
\newblock \bibinfo{title}{Artificial emotional creature for human-machine interaction},
\newblock in: \bibinfo{booktitle}{1997 IEEE international conference on systems, man, and cybernetics. Computational cybernetics and simulation}, volume~\bibinfo{volume}{3}, \bibinfo{organization}{IEEE}, \bibinfo{year}{1997}, pp. \bibinfo{pages}{2269--2274}.
\bibitem[{Sun et~al.(2019)Sun, Pei, Zhang, Li, and Tao}]{sun2019design}
\bibinfo{author}{X.~Sun}, \bibinfo{author}{Z.~Pei}, \bibinfo{author}{C.~Zhang}, \bibinfo{author}{G.~Li}, \bibinfo{author}{J.~Tao},
\newblock \bibinfo{title}{Design and analysis of a human--machine interaction system for researching human’s dynamic emotion},
\newblock \bibinfo{journal}{IEEE Transactions on Systems, Man, and Cybernetics: Systems} \bibinfo{volume}{51} (\bibinfo{year}{2019}) \bibinfo{pages}{6111--6121}.
\bibitem[{Erol et~al.(2019)Erol, Majumdar, Benavidez, Rad, Choo, and Jamshidi}]{erol2019toward}
\bibinfo{author}{B.~A. Erol}, \bibinfo{author}{A.~Majumdar}, \bibinfo{author}{P.~Benavidez}, \bibinfo{author}{P.~Rad}, \bibinfo{author}{K.-K.~R. Choo}, \bibinfo{author}{M.~Jamshidi},
\newblock \bibinfo{title}{Toward artificial emotional intelligence for cooperative social human--machine interaction},
\newblock \bibinfo{journal}{IEEE Transactions on Computational Social Systems} \bibinfo{volume}{7} (\bibinfo{year}{2019}) \bibinfo{pages}{234--246}.
\bibitem[{Volonte et~al.(2021)Volonte, Wang, Ebrahimi, Hsu, Liu, Wong, and Babu}]{volonte2021effects}
\bibinfo{author}{M.~Volonte}, \bibinfo{author}{C.-C. Wang}, \bibinfo{author}{E.~Ebrahimi}, \bibinfo{author}{Y.-C. Hsu}, \bibinfo{author}{K.-Y. Liu}, \bibinfo{author}{S.-K. Wong}, \bibinfo{author}{S.~V. Babu},
\newblock \bibinfo{title}{Effects of language familiarity in simulated natural dialogue with a virtual crowd of digital humans on emotion contagion in virtual reality},
\newblock in: \bibinfo{booktitle}{2021 IEEE Virtual Reality and 3D User Interfaces (VR)}, \bibinfo{organization}{IEEE}, \bibinfo{year}{2021}, pp. \bibinfo{pages}{188--197}.
\bibitem[{Loveys et~al.(2021)Loveys, Sagar, Zhang, Fricchione, and Broadbent}]{loveys2021effects}
\bibinfo{author}{K.~Loveys}, \bibinfo{author}{M.~Sagar}, \bibinfo{author}{X.~Zhang}, \bibinfo{author}{G.~Fricchione}, \bibinfo{author}{E.~Broadbent},
\newblock \bibinfo{title}{Effects of emotional expressiveness of a female digital human on loneliness, stress, perceived support, and closeness across genders: randomized controlled trial},
\newblock \bibinfo{journal}{Journal of medical Internet research} \bibinfo{volume}{23} (\bibinfo{year}{2021}) \bibinfo{pages}{e30624}.
\bibitem[{Li et~al.(2017)Li, Deng, and Du}]{li2017reliable}
\bibinfo{author}{S.~Li}, \bibinfo{author}{W.~Deng}, \bibinfo{author}{J.~Du},
\newblock \bibinfo{title}{Reliable crowdsourcing and deep locality-preserving learning for expression recognition in the wild},
\newblock in: \bibinfo{booktitle}{Proceedings of the IEEE conference on computer vision and pattern recognition}, \bibinfo{year}{2017}, pp. \bibinfo{pages}{2852--2861}.
\bibitem[{Li et~al.(2021)Li, Wang, Ding, Yang, and Gao}]{li2021adaptively}
\bibinfo{author}{H.~Li}, \bibinfo{author}{N.~Wang}, \bibinfo{author}{X.~Ding}, \bibinfo{author}{X.~Yang}, \bibinfo{author}{X.~Gao},
\newblock \bibinfo{title}{Adaptively learning facial expression representation via cf labels and distillation},
\newblock \bibinfo{journal}{IEEE Transactions on Image Processing} \bibinfo{volume}{30} (\bibinfo{year}{2021}) \bibinfo{pages}{2016--2028}.
\bibitem[{She et~al.(2021)She, Hu, Shi, Wang, Shen, and Mei}]{she2021dive}
\bibinfo{author}{J.~She}, \bibinfo{author}{Y.~Hu}, \bibinfo{author}{H.~Shi}, \bibinfo{author}{J.~Wang}, \bibinfo{author}{Q.~Shen}, \bibinfo{author}{T.~Mei},
\newblock \bibinfo{title}{Dive into ambiguity: Latent distribution mining and pairwise uncertainty estimation for facial expression recognition},
\newblock in: \bibinfo{booktitle}{Proceedings of the IEEE/CVF conference on computer vision and pattern recognition}, \bibinfo{year}{2021}, pp. \bibinfo{pages}{6248--6257}.
\bibitem[{Xue et~al.(2021)Xue, Wang, and Guo}]{xue2021transfer}
\bibinfo{author}{F.~Xue}, \bibinfo{author}{Q.~Wang}, \bibinfo{author}{G.~Guo},
\newblock \bibinfo{title}{Transfer: Learning relation-aware facial expression representations with transformers},
\newblock in: \bibinfo{booktitle}{Proceedings of the IEEE/CVF International Conference on Computer Vision}, \bibinfo{year}{2021}, pp. \bibinfo{pages}{3601--3610}.
\bibitem[{Grandvalet and Bengio(2004)}]{grandvalet2004semi}
\bibinfo{author}{Y.~Grandvalet}, \bibinfo{author}{Y.~Bengio},
\newblock \bibinfo{title}{Semi-supervised learning by entropy minimization},
\newblock \bibinfo{journal}{Advances in neural information processing systems} \bibinfo{volume}{17} (\bibinfo{year}{2004}).
\bibitem[{Rosenberg et~al.(2005)Rosenberg, Hebert, and Schneiderman}]{rosenberg2005semi}
\bibinfo{author}{C.~Rosenberg}, \bibinfo{author}{M.~Hebert}, \bibinfo{author}{H.~Schneiderman},
\newblock \bibinfo{title}{Semi-supervised self-training of object detection models}  (\bibinfo{year}{2005}).
\bibitem[{Hadsell et~al.(2006)Hadsell, Chopra, and LeCun}]{hadsell2006dimensionality}
\bibinfo{author}{R.~Hadsell}, \bibinfo{author}{S.~Chopra}, \bibinfo{author}{Y.~LeCun},
\newblock \bibinfo{title}{Dimensionality reduction by learning an invariant mapping},
\newblock in: \bibinfo{booktitle}{2006 IEEE Computer Society Conference on Computer Vision and Pattern Recognition (CVPR'06)}, volume~\bibinfo{volume}{2}, \bibinfo{organization}{IEEE}, \bibinfo{year}{2006}, pp. \bibinfo{pages}{1735--1742}.
\bibitem[{Song et~al.(2017)Song, Yu, Zeng, Chang, Savva, and Funkhouser}]{song2017semantic}
\bibinfo{author}{S.~Song}, \bibinfo{author}{F.~Yu}, \bibinfo{author}{A.~Zeng}, \bibinfo{author}{A.~X. Chang}, \bibinfo{author}{M.~Savva}, \bibinfo{author}{T.~Funkhouser},
\newblock \bibinfo{title}{Semantic scene completion from a single depth image},
\newblock in: \bibinfo{booktitle}{Proceedings of the IEEE conference on computer vision and pattern recognition}, \bibinfo{year}{2017}, pp. \bibinfo{pages}{1746--1754}.
\bibitem[{Li et~al.(2022)Li, Wang, Yang, Wang, and Gao}]{adacm}
\bibinfo{author}{H.~Li}, \bibinfo{author}{N.~Wang}, \bibinfo{author}{X.~Yang}, \bibinfo{author}{X.~Wang}, \bibinfo{author}{X.~Gao},
\newblock \bibinfo{title}{Towards semi-supervised deep facial expression recognition with an adaptive confidence margin},
\newblock in: \bibinfo{booktitle}{Proceedings of the IEEE/CVF conference on computer vision and pattern recognition}, \bibinfo{year}{2022}, pp. \bibinfo{pages}{4166--4175}.
\bibitem[{Florea et~al.(2020)Florea, Badea, Florea, Racoviteanu, and Vertan}]{margin-mix}
\bibinfo{author}{C.~Florea}, \bibinfo{author}{M.~Badea}, \bibinfo{author}{L.~Florea}, \bibinfo{author}{A.~Racoviteanu}, \bibinfo{author}{C.~Vertan},
\newblock \bibinfo{title}{Margin-mix: Semi-supervised learning for face expression recognition},
\newblock in: \bibinfo{booktitle}{Computer Vision--ECCV 2020: 16th European Conference, Glasgow, UK, August 23--28, 2020, Proceedings, Part XXIII 16}, \bibinfo{organization}{Springer}, \bibinfo{year}{2020}, pp. \bibinfo{pages}{1--17}.
\bibitem[{Sohn et~al.(2020)Sohn, Berthelot, Carlini, Zhang, Zhang, Raffel, Cubuk, Kurakin, and Li}]{fixmatch}
\bibinfo{author}{K.~Sohn}, \bibinfo{author}{D.~Berthelot}, \bibinfo{author}{N.~Carlini}, \bibinfo{author}{Z.~Zhang}, \bibinfo{author}{H.~Zhang}, \bibinfo{author}{C.~A. Raffel}, \bibinfo{author}{E.~D. Cubuk}, \bibinfo{author}{A.~Kurakin}, \bibinfo{author}{C.-L. Li},
\newblock \bibinfo{title}{Fixmatch: Simplifying semi-supervised learning with consistency and confidence},
\newblock \bibinfo{journal}{Advances in neural information processing systems} \bibinfo{volume}{33} (\bibinfo{year}{2020}) \bibinfo{pages}{596--608}.
\bibitem[{Zhang et~al.(2021)Zhang, Wang, Hou, Wu, Wang, Okumura, and Shinozaki}]{flexmatch}
\bibinfo{author}{B.~Zhang}, \bibinfo{author}{Y.~Wang}, \bibinfo{author}{W.~Hou}, \bibinfo{author}{H.~Wu}, \bibinfo{author}{J.~Wang}, \bibinfo{author}{M.~Okumura}, \bibinfo{author}{T.~Shinozaki},
\newblock \bibinfo{title}{Flexmatch: Boosting semi-supervised learning with curriculum pseudo labeling},
\newblock \bibinfo{journal}{Advances in Neural Information Processing Systems} \bibinfo{volume}{34} (\bibinfo{year}{2021}) \bibinfo{pages}{18408--18419}.
\bibitem[{Wang et~al.(2022)Wang, Chen, Heng, Hou, Fan, Wu, Wang, Savvides, Shinozaki, Raj et~al.}]{wang2022freematch}
\bibinfo{author}{Y.~Wang}, \bibinfo{author}{H.~Chen}, \bibinfo{author}{Q.~Heng}, \bibinfo{author}{W.~Hou}, \bibinfo{author}{Y.~Fan}, \bibinfo{author}{Z.~Wu}, \bibinfo{author}{J.~Wang}, \bibinfo{author}{M.~Savvides}, \bibinfo{author}{T.~Shinozaki}, \bibinfo{author}{B.~Raj}, et~al.,
\newblock \bibinfo{title}{Freematch: Self-adaptive thresholding for semi-supervised learning},
\newblock \bibinfo{journal}{arXiv preprint arXiv:2205.07246}  (\bibinfo{year}{2022}).
\bibitem[{Chen et~al.(2023)Chen, Tao, Fan, Wang, Wang, Schiele, Xie, Raj, and Savvides}]{chen2023softmatch}
\bibinfo{author}{H.~Chen}, \bibinfo{author}{R.~Tao}, \bibinfo{author}{Y.~Fan}, \bibinfo{author}{Y.~Wang}, \bibinfo{author}{J.~Wang}, \bibinfo{author}{B.~Schiele}, \bibinfo{author}{X.~Xie}, \bibinfo{author}{B.~Raj}, \bibinfo{author}{M.~Savvides},
\newblock \bibinfo{title}{Softmatch: Addressing the quantity-quality trade-off in semi-supervised learning},
\newblock \bibinfo{journal}{arXiv preprint arXiv:2301.10921}  (\bibinfo{year}{2023}).
\bibitem[{Eigen et~al.(2013)Eigen, Ranzato, and Sutskever}]{eigen2013learning}
\bibinfo{author}{D.~Eigen}, \bibinfo{author}{M.~Ranzato}, \bibinfo{author}{I.~Sutskever},
\newblock \bibinfo{title}{Learning factored representations in a deep mixture of experts},
\newblock \bibinfo{journal}{arXiv preprint arXiv:1312.4314}  (\bibinfo{year}{2013}).
\bibitem[{Shazeer et~al.(2017)Shazeer, Mirhoseini, Maziarz, Davis, Le, Hinton, and Dean}]{shazeer2017outrageously}
\bibinfo{author}{N.~Shazeer}, \bibinfo{author}{A.~Mirhoseini}, \bibinfo{author}{K.~Maziarz}, \bibinfo{author}{A.~Davis}, \bibinfo{author}{Q.~Le}, \bibinfo{author}{G.~Hinton}, \bibinfo{author}{J.~Dean},
\newblock \bibinfo{title}{Outrageously large neural networks: The sparsely-gated mixture-of-experts layer},
\newblock \bibinfo{journal}{arXiv preprint arXiv:1701.06538}  (\bibinfo{year}{2017}).
\bibitem[{Fedus et~al.(2022)Fedus, Zoph, and Shazeer}]{fedus2022switch}
\bibinfo{author}{W.~Fedus}, \bibinfo{author}{B.~Zoph}, \bibinfo{author}{N.~Shazeer},
\newblock \bibinfo{title}{Switch transformers: Scaling to trillion parameter models with simple and efficient sparsity},
\newblock \bibinfo{journal}{The Journal of Machine Learning Research} \bibinfo{volume}{23} (\bibinfo{year}{2022}) \bibinfo{pages}{5232--5270}.
\bibitem[{Liu et~al.(2017)Liu, Wen, Yu, Li, Raj, and Song}]{liu2017sphereface}
\bibinfo{author}{W.~Liu}, \bibinfo{author}{Y.~Wen}, \bibinfo{author}{Z.~Yu}, \bibinfo{author}{M.~Li}, \bibinfo{author}{B.~Raj}, \bibinfo{author}{L.~Song},
\newblock \bibinfo{title}{Sphereface: Deep hypersphere embedding for face recognition},
\newblock in: \bibinfo{booktitle}{Proceedings of the IEEE conference on computer vision and pattern recognition}, \bibinfo{year}{2017}, pp. \bibinfo{pages}{212--220}.
\bibitem[{Sun et~al.(2020)Sun, Cheng, Zhang, Zhang, Zheng, Wang, and Wei}]{sun2020circle}
\bibinfo{author}{Y.~Sun}, \bibinfo{author}{C.~Cheng}, \bibinfo{author}{Y.~Zhang}, \bibinfo{author}{C.~Zhang}, \bibinfo{author}{L.~Zheng}, \bibinfo{author}{Z.~Wang}, \bibinfo{author}{Y.~Wei},
\newblock \bibinfo{title}{Circle loss: A unified perspective of pair similarity optimization},
\newblock in: \bibinfo{booktitle}{Proceedings of the IEEE/CVF conference on computer vision and pattern recognition}, \bibinfo{year}{2020}, pp. \bibinfo{pages}{6398--6407}.
\bibitem[{Wang et~al.(2018{\natexlab{a}})Wang, Cheng, Liu, and Liu}]{wang2018additive}
\bibinfo{author}{F.~Wang}, \bibinfo{author}{J.~Cheng}, \bibinfo{author}{W.~Liu}, \bibinfo{author}{H.~Liu},
\newblock \bibinfo{title}{Additive margin softmax for face verification},
\newblock \bibinfo{journal}{IEEE Signal Processing Letters} \bibinfo{volume}{25} (\bibinfo{year}{2018}{\natexlab{a}}) \bibinfo{pages}{926--930}.
\bibitem[{Wang et~al.(2018{\natexlab{b}})Wang, Wang, Zhou, Ji, Gong, Zhou, Li, and Liu}]{wang2018cosface}
\bibinfo{author}{H.~Wang}, \bibinfo{author}{Y.~Wang}, \bibinfo{author}{Z.~Zhou}, \bibinfo{author}{X.~Ji}, \bibinfo{author}{D.~Gong}, \bibinfo{author}{J.~Zhou}, \bibinfo{author}{Z.~Li}, \bibinfo{author}{W.~Liu},
\newblock \bibinfo{title}{Cosface: Large margin cosine loss for deep face recognition},
\newblock in: \bibinfo{booktitle}{Proceedings of the IEEE conference on computer vision and pattern recognition}, \bibinfo{year}{2018}{\natexlab{b}}, pp. \bibinfo{pages}{5265--5274}.
\bibitem[{Qiao et~al.(2023)Qiao, Wei, Wang, Wang, Song, Xu, Ji, Liu, and Chen}]{fpl}
\bibinfo{author}{P.~Qiao}, \bibinfo{author}{Z.~Wei}, \bibinfo{author}{Y.~Wang}, \bibinfo{author}{Z.~Wang}, \bibinfo{author}{G.~Song}, \bibinfo{author}{F.~Xu}, \bibinfo{author}{X.~Ji}, \bibinfo{author}{C.~Liu}, \bibinfo{author}{J.~Chen},
\newblock \bibinfo{title}{Fuzzy positive learning for semi-supervised semantic segmentation},
\newblock in: \bibinfo{booktitle}{Proceedings of the IEEE/CVF Conference on Computer Vision and Pattern Recognition}, \bibinfo{year}{2023}, pp. \bibinfo{pages}{15465--15474}.
\bibitem[{Hu et~al.(2008)Hu, Zeng, Yin, Wei, Zhou, and Huang}]{hu2008multi}
\bibinfo{author}{Y.~Hu}, \bibinfo{author}{Z.~Zeng}, \bibinfo{author}{L.~Yin}, \bibinfo{author}{X.~Wei}, \bibinfo{author}{X.~Zhou}, \bibinfo{author}{T.~S. Huang},
\newblock \bibinfo{title}{Multi-view facial expression recognition},
\newblock in: \bibinfo{booktitle}{2008 8th IEEE International Conference on Automatic Face \& Gesture Recognition}, \bibinfo{organization}{IEEE}, \bibinfo{year}{2008}, pp. \bibinfo{pages}{1--6}.
\bibitem[{Luo et~al.(2013)Luo, Wu, and Zhang}]{luo2013facial}
\bibinfo{author}{Y.~Luo}, \bibinfo{author}{C.-m. Wu}, \bibinfo{author}{Y.~Zhang},
\newblock \bibinfo{title}{Facial expression recognition based on fusion feature of pca and lbp with svm},
\newblock \bibinfo{journal}{Optik-International Journal for Light and Electron Optics} \bibinfo{volume}{124} (\bibinfo{year}{2013}) \bibinfo{pages}{2767--2770}.
\bibitem[{Pietik{\"a}inen et~al.(2011)Pietik{\"a}inen, Hadid, Zhao, and Ahonen}]{pietikainen2011computer}
\bibinfo{author}{M.~Pietik{\"a}inen}, \bibinfo{author}{A.~Hadid}, \bibinfo{author}{G.~Zhao}, \bibinfo{author}{T.~Ahonen}, \bibinfo{title}{Computer vision using local binary patterns}, volume~\bibinfo{volume}{40}, \bibinfo{publisher}{Springer Science \& Business Media}, \bibinfo{year}{2011}.
\bibitem[{Lucey et~al.(2010)Lucey, Cohn, Kanade, Saragih, Ambadar, and Matthews}]{lucey2010extended}
\bibinfo{author}{P.~Lucey}, \bibinfo{author}{J.~F. Cohn}, \bibinfo{author}{T.~Kanade}, \bibinfo{author}{J.~Saragih}, \bibinfo{author}{Z.~Ambadar}, \bibinfo{author}{I.~Matthews},
\newblock \bibinfo{title}{The extended cohn-kanade dataset (ck+): A complete dataset for action unit and emotion-specified expression},
\newblock in: \bibinfo{booktitle}{2010 ieee computer society conference on computer vision and pattern recognition-workshops}, \bibinfo{organization}{IEEE}, \bibinfo{year}{2010}, pp. \bibinfo{pages}{94--101}.
\bibitem[{Zhao et~al.(2011)Zhao, Huang, Taini, Li, and Pietik{\"a}Inen}]{zhao2011facial}
\bibinfo{author}{G.~Zhao}, \bibinfo{author}{X.~Huang}, \bibinfo{author}{M.~Taini}, \bibinfo{author}{S.~Z. Li}, \bibinfo{author}{M.~Pietik{\"a}Inen},
\newblock \bibinfo{title}{Facial expression recognition from near-infrared videos},
\newblock \bibinfo{journal}{Image and vision computing} \bibinfo{volume}{29} (\bibinfo{year}{2011}) \bibinfo{pages}{607--619}.
\bibitem[{Barsoum et~al.(2016)Barsoum, Zhang, Ferrer, and Zhang}]{barsoum2016training}
\bibinfo{author}{E.~Barsoum}, \bibinfo{author}{C.~Zhang}, \bibinfo{author}{C.~C. Ferrer}, \bibinfo{author}{Z.~Zhang},
\newblock \bibinfo{title}{Training deep networks for facial expression recognition with crowd-sourced label distribution},
\newblock in: \bibinfo{booktitle}{Proceedings of the 18th ACM international conference on multimodal interaction}, \bibinfo{year}{2016}, pp. \bibinfo{pages}{279--283}.
\bibitem[{Mollahosseini et~al.(2017)Mollahosseini, Hasani, and Mahoor}]{mollahosseini2017affectnet}
\bibinfo{author}{A.~Mollahosseini}, \bibinfo{author}{B.~Hasani}, \bibinfo{author}{M.~H. Mahoor},
\newblock \bibinfo{title}{Affectnet: A database for facial expression, valence, and arousal computing in the wild},
\newblock \bibinfo{journal}{IEEE Transactions on Affective Computing} \bibinfo{volume}{10} (\bibinfo{year}{2017}) \bibinfo{pages}{18--31}.
\bibitem[{Zhang et~al.(2024)Zhang, Li, Liu, Deng et~al.}]{zhang2024leave}
\bibinfo{author}{Y.~Zhang}, \bibinfo{author}{Y.~Li}, \bibinfo{author}{X.~Liu}, \bibinfo{author}{W.~Deng}, et~al.,
\newblock \bibinfo{title}{Leave no stone unturned: Mine extra knowledge for imbalanced facial expression recognition},
\newblock \bibinfo{journal}{Advances in Neural Information Processing Systems} \bibinfo{volume}{36} (\bibinfo{year}{2024}).
\bibitem[{Wu and Cui(2023)}]{wu2023net}
\bibinfo{author}{Z.~Wu}, \bibinfo{author}{J.~Cui},
\newblock \bibinfo{title}{La-net: Landmark-aware learning for reliable facial expression recognition under label noise},
\newblock in: \bibinfo{booktitle}{Proceedings of the IEEE/CVF International Conference on Computer Vision}, \bibinfo{year}{2023}, pp. \bibinfo{pages}{20698--20707}.
\bibitem[{Chen et~al.(2024)Chen, Wen, Yang, Li, Chen, and Wang}]{chen2024cfan}
\bibinfo{author}{D.~Chen}, \bibinfo{author}{G.~Wen}, \bibinfo{author}{P.~Yang}, \bibinfo{author}{H.~Li}, \bibinfo{author}{C.~Chen}, \bibinfo{author}{B.~Wang},
\newblock \bibinfo{title}{Cfan-sda: Coarse-fine aware network with static-dynamic adaptation for facial expression recognition in videos},
\newblock \bibinfo{journal}{IEEE Transactions on Circuits and Systems for Video Technology}  (\bibinfo{year}{2024}).
\bibitem[{Xie et~al.(2020)Xie, Hu, and Chen}]{xie2020facial}
\bibinfo{author}{S.~Xie}, \bibinfo{author}{H.~Hu}, \bibinfo{author}{Y.~Chen},
\newblock \bibinfo{title}{Facial expression recognition with two-branch disentangled generative adversarial network},
\newblock \bibinfo{journal}{IEEE Transactions on Circuits and Systems for Video Technology} \bibinfo{volume}{31} (\bibinfo{year}{2020}) \bibinfo{pages}{2359--2371}.
\bibitem[{Wang et~al.(2021)Wang, Xue, Lu, and Yan}]{wang2021light}
\bibinfo{author}{C.~Wang}, \bibinfo{author}{J.~Xue}, \bibinfo{author}{K.~Lu}, \bibinfo{author}{Y.~Yan},
\newblock \bibinfo{title}{Light attention embedding for facial expression recognition},
\newblock \bibinfo{journal}{IEEE Transactions on Circuits and Systems for Video Technology} \bibinfo{volume}{32} (\bibinfo{year}{2021}) \bibinfo{pages}{1834--1847}.
\bibitem[{Li et~al.(2021)Li, Lu, Chen, Zhang, Li, Lu, and Zhang}]{li2021learning}
\bibinfo{author}{Y.~Li}, \bibinfo{author}{Y.~Lu}, \bibinfo{author}{B.~Chen}, \bibinfo{author}{Z.~Zhang}, \bibinfo{author}{J.~Li}, \bibinfo{author}{G.~Lu}, \bibinfo{author}{D.~Zhang},
\newblock \bibinfo{title}{Learning informative and discriminative features for facial expression recognition in the wild},
\newblock \bibinfo{journal}{IEEE Transactions on Circuits and Systems for Video Technology} \bibinfo{volume}{32} (\bibinfo{year}{2021}) \bibinfo{pages}{3178--3189}.
\bibitem[{Gu et~al.(2022)Gu, Yan, Zhang, Wang, Ji, and Ren}]{gu2022toward}
\bibinfo{author}{Y.~Gu}, \bibinfo{author}{H.~Yan}, \bibinfo{author}{X.~Zhang}, \bibinfo{author}{Y.~Wang}, \bibinfo{author}{Y.~Ji}, \bibinfo{author}{F.~Ren},
\newblock \bibinfo{title}{Toward facial expression recognition in the wild via noise-tolerant network},
\newblock \bibinfo{journal}{IEEE Transactions on Circuits and Systems for Video Technology} \bibinfo{volume}{33} (\bibinfo{year}{2022}) \bibinfo{pages}{2033--2047}.
\bibitem[{Li et~al.(2023)Li, Li, Wang, Huang, Liu, and Liao}]{li2023fg}
\bibinfo{author}{C.~Li}, \bibinfo{author}{X.~Li}, \bibinfo{author}{X.~Wang}, \bibinfo{author}{D.~Huang}, \bibinfo{author}{Z.~Liu}, \bibinfo{author}{L.~Liao},
\newblock \bibinfo{title}{Fg-agr: Fine-grained associative graph representation for facial expression recognition in the wild},
\newblock \bibinfo{journal}{IEEE Transactions on Circuits and Systems for Video Technology} \bibinfo{volume}{34} (\bibinfo{year}{2023}) \bibinfo{pages}{882--896}.
\bibitem[{Cai et~al.(2024)Cai, Zhao, Yi, Yu, Duan, Pan, and Liu}]{cai2024mfdan}
\bibinfo{author}{W.~Cai}, \bibinfo{author}{J.~Zhao}, \bibinfo{author}{R.~Yi}, \bibinfo{author}{M.~Yu}, \bibinfo{author}{F.~Duan}, \bibinfo{author}{Z.~Pan}, \bibinfo{author}{Y.-J. Liu},
\newblock \bibinfo{title}{Mfdan: Multi-level flow-driven attention network for micro-expression recognition},
\newblock \bibinfo{journal}{IEEE Transactions on Circuits and Systems for Video Technology}  (\bibinfo{year}{2024}).
\bibitem[{Liu et~al.(2022)Liu, Jiang, Li, Guo, Jiang, and Ren}]{liu2022devil}
\bibinfo{author}{H.~Liu}, \bibinfo{author}{X.~Jiang}, \bibinfo{author}{X.~Li}, \bibinfo{author}{A.~Guo}, \bibinfo{author}{D.~Jiang}, \bibinfo{author}{B.~Ren}, \bibinfo{title}{The devil is in the frequency: Geminated gestalt autoencoder for self-supervised visual pre-training}, \bibinfo{year}{2022}. \href{http://arxiv.org/abs/2204.08227}{\tt arXiv:2204.08227}.
\bibitem[{Zhang et~al.(2024)Zhang, Jia, Wang, Che, and Sun}]{zhang2024self}
\bibinfo{author}{W.-L. Zhang}, \bibinfo{author}{R.-S. Jia}, \bibinfo{author}{H.~Wang}, \bibinfo{author}{C.-Y. Che}, \bibinfo{author}{H.-M. Sun},
\newblock \bibinfo{title}{A self-supervised learning network for student engagement recognition from facial expressions},
\newblock \bibinfo{journal}{IEEE Transactions on Circuits and Systems for Video Technology}  (\bibinfo{year}{2024}).
\bibitem[{Zhou et~al.(2024)Zhou, Huang, Zhang, and Xu}]{zhou2024ceprompt}
\bibinfo{author}{H.~Zhou}, \bibinfo{author}{S.~Huang}, \bibinfo{author}{F.~Zhang}, \bibinfo{author}{C.~Xu},
\newblock \bibinfo{title}{Ceprompt: Cross-modal emotion-aware prompting for facial expression recognition},
\newblock \bibinfo{journal}{IEEE Transactions on Circuits and Systems for Video Technology}  (\bibinfo{year}{2024}).
\bibitem[{Sajjadi et~al.(2016)Sajjadi, Javanmardi, and Tasdizen}]{regularization}
\bibinfo{author}{M.~Sajjadi}, \bibinfo{author}{M.~Javanmardi}, \bibinfo{author}{T.~Tasdizen},
\newblock \bibinfo{title}{Regularization with stochastic transformations and perturbations for deep semi-supervised learning},
\newblock \bibinfo{journal}{Advances in neural information processing systems} \bibinfo{volume}{29} (\bibinfo{year}{2016}).
\bibitem[{Xie et~al.(2020)Xie, Dai, Hovy, Luong, and Le}]{uda}
\bibinfo{author}{Q.~Xie}, \bibinfo{author}{Z.~Dai}, \bibinfo{author}{E.~Hovy}, \bibinfo{author}{T.~Luong}, \bibinfo{author}{Q.~Le},
\newblock \bibinfo{title}{Unsupervised data augmentation for consistency training},
\newblock \bibinfo{journal}{Advances in neural information processing systems} \bibinfo{volume}{33} (\bibinfo{year}{2020}) \bibinfo{pages}{6256--6268}.
\bibitem[{Lee et~al.(2013)}]{lee2013pseudo}
\bibinfo{author}{D.-H. Lee}, et~al.,
\newblock \bibinfo{title}{Pseudo-label: The simple and efficient semi-supervised learning method for deep neural networks},
\newblock in: \bibinfo{booktitle}{Workshop on challenges in representation learning, ICML}, volume~\bibinfo{volume}{3}, \bibinfo{organization}{Atlanta}, \bibinfo{year}{2013}, p. \bibinfo{pages}{896}.
\bibitem[{Berthelot et~al.(2019)Berthelot, Carlini, Goodfellow, Papernot, Oliver, and Raffel}]{berthelot2019mixmatch}
\bibinfo{author}{D.~Berthelot}, \bibinfo{author}{N.~Carlini}, \bibinfo{author}{I.~Goodfellow}, \bibinfo{author}{N.~Papernot}, \bibinfo{author}{A.~Oliver}, \bibinfo{author}{C.~A. Raffel},
\newblock \bibinfo{title}{Mixmatch: A holistic approach to semi-supervised learning},
\newblock \bibinfo{journal}{Advances in neural information processing systems} \bibinfo{volume}{32} (\bibinfo{year}{2019}).
\bibitem[{Xu et~al.(2021)Xu, Shang, Ye, Qian, Li, Sun, Li, and Jin}]{xu2021dash}
\bibinfo{author}{Y.~Xu}, \bibinfo{author}{L.~Shang}, \bibinfo{author}{J.~Ye}, \bibinfo{author}{Q.~Qian}, \bibinfo{author}{Y.-F. Li}, \bibinfo{author}{B.~Sun}, \bibinfo{author}{H.~Li}, \bibinfo{author}{R.~Jin},
\newblock \bibinfo{title}{Dash: Semi-supervised learning with dynamic thresholding},
\newblock in: \bibinfo{booktitle}{International Conference on Machine Learning}, \bibinfo{organization}{PMLR}, \bibinfo{year}{2021}, pp. \bibinfo{pages}{11525--11536}.
\bibitem[{Laine and Aila(2016)}]{laine2016temporal}
\bibinfo{author}{S.~Laine}, \bibinfo{author}{T.~Aila},
\newblock \bibinfo{title}{Temporal ensembling for semi-supervised learning},
\newblock \bibinfo{journal}{arXiv preprint arXiv:1610.02242}  (\bibinfo{year}{2016}).
\bibitem[{Miyato et~al.(2018)Miyato, Maeda, Koyama, and Ishii}]{miyato2018virtual}
\bibinfo{author}{T.~Miyato}, \bibinfo{author}{S.-i. Maeda}, \bibinfo{author}{M.~Koyama}, \bibinfo{author}{S.~Ishii},
\newblock \bibinfo{title}{Virtual adversarial training: a regularization method for supervised and semi-supervised learning},
\newblock \bibinfo{journal}{IEEE transactions on pattern analysis and machine intelligence} \bibinfo{volume}{41} (\bibinfo{year}{2018}) \bibinfo{pages}{1979--1993}.
\bibitem[{Tarvainen and Valpola(2017)}]{tarvainen2017mean}
\bibinfo{author}{A.~Tarvainen}, \bibinfo{author}{H.~Valpola},
\newblock \bibinfo{title}{Mean teachers are better role models: Weight-averaged consistency targets improve semi-supervised deep learning results},
\newblock \bibinfo{journal}{Advances in neural information processing systems} \bibinfo{volume}{30} (\bibinfo{year}{2017}).
\bibitem[{Berthelot et~al.(2019)Berthelot, Carlini, Cubuk, Kurakin, Sohn, Zhang, and Raffel}]{berthelot2019remixmatch}
\bibinfo{author}{D.~Berthelot}, \bibinfo{author}{N.~Carlini}, \bibinfo{author}{E.~D. Cubuk}, \bibinfo{author}{A.~Kurakin}, \bibinfo{author}{K.~Sohn}, \bibinfo{author}{H.~Zhang}, \bibinfo{author}{C.~Raffel},
\newblock \bibinfo{title}{Remixmatch: Semi-supervised learning with distribution alignment and augmentation anchoring},
\newblock \bibinfo{journal}{arXiv preprint arXiv:1911.09785}  (\bibinfo{year}{2019}).
\bibitem[{Schmutz et~al.(2022)Schmutz, Humbert, and Mattei}]{schmutz2022don}
\bibinfo{author}{H.~Schmutz}, \bibinfo{author}{O.~Humbert}, \bibinfo{author}{P.-A. Mattei},
\newblock \bibinfo{title}{Don’t fear the unlabelled: safe semi-supervised learning via debiasing},
\newblock in: \bibinfo{booktitle}{The Eleventh International Conference on Learning Representations}, \bibinfo{year}{2022}.
\bibitem[{Li et~al.(2021)Li, Xiong, and Hoi}]{li2021comatch}
\bibinfo{author}{J.~Li}, \bibinfo{author}{C.~Xiong}, \bibinfo{author}{S.~C. Hoi},
\newblock \bibinfo{title}{Comatch: Semi-supervised learning with contrastive graph regularization},
\newblock in: \bibinfo{booktitle}{Proceedings of the IEEE/CVF International Conference on Computer Vision}, \bibinfo{year}{2021}, pp. \bibinfo{pages}{9475--9484}.
\bibitem[{Zheng et~al.(2022)Zheng, You, Huang, Wang, Qian, and Xu}]{zheng2022simmatch}
\bibinfo{author}{M.~Zheng}, \bibinfo{author}{S.~You}, \bibinfo{author}{L.~Huang}, \bibinfo{author}{F.~Wang}, \bibinfo{author}{C.~Qian}, \bibinfo{author}{C.~Xu},
\newblock \bibinfo{title}{Simmatch: Semi-supervised learning with similarity matching},
\newblock in: \bibinfo{booktitle}{Proceedings of the IEEE/CVF Conference on Computer Vision and Pattern Recognition}, \bibinfo{year}{2022}, pp. \bibinfo{pages}{14471--14481}.
\bibitem[{Berthelot et~al.(2021)Berthelot, Roelofs, Sohn, Carlini, and Kurakin}]{berthelot2021adamatch}
\bibinfo{author}{D.~Berthelot}, \bibinfo{author}{R.~Roelofs}, \bibinfo{author}{K.~Sohn}, \bibinfo{author}{N.~Carlini}, \bibinfo{author}{A.~Kurakin},
\newblock \bibinfo{title}{Adamatch: A unified approach to semi-supervised learning and domain adaptation},
\newblock \bibinfo{journal}{arXiv preprint arXiv:2106.04732}  (\bibinfo{year}{2021}).
\bibitem[{Du et~al.(2023)Du, Jiang, Wang, Zhou, Wu, Zhou, and Wang}]{du2023lion}
\bibinfo{author}{Z.~Du}, \bibinfo{author}{X.~Jiang}, \bibinfo{author}{P.~Wang}, \bibinfo{author}{Q.~Zhou}, \bibinfo{author}{X.~Wu}, \bibinfo{author}{J.~Zhou}, \bibinfo{author}{Y.~Wang},
\newblock \bibinfo{title}{Lion: label disambiguation for semi-supervised facial expression recognition with progressive negative learning},
\newblock in: \bibinfo{booktitle}{Proceedings of the Thirty-Second International Joint Conference on Artificial Intelligence}, \bibinfo{year}{2023}, pp. \bibinfo{pages}{699--707}.
\bibitem[{He et~al.(2016)He, Zhang, Ren, and Sun}]{he2016deep}
\bibinfo{author}{K.~He}, \bibinfo{author}{X.~Zhang}, \bibinfo{author}{S.~Ren}, \bibinfo{author}{J.~Sun},
\newblock \bibinfo{title}{Deep residual learning for image recognition},
\newblock in: \bibinfo{booktitle}{Proceedings of the IEEE conference on computer vision and pattern recognition}, \bibinfo{year}{2016}, pp. \bibinfo{pages}{770--778}.
\bibitem[{Guo et~al.(2016)Guo, Zhang, Hu, He, and Gao}]{guo2016ms}
\bibinfo{author}{Y.~Guo}, \bibinfo{author}{L.~Zhang}, \bibinfo{author}{Y.~Hu}, \bibinfo{author}{X.~He}, \bibinfo{author}{J.~Gao},
\newblock \bibinfo{title}{Ms-celeb-1m: A dataset and benchmark for large-scale face recognition},
\newblock in: \bibinfo{booktitle}{Computer Vision--ECCV 2016: 14th European Conference, Amsterdam, The Netherlands, October 11-14, 2016, Proceedings, Part III 14}, \bibinfo{organization}{Springer}, \bibinfo{year}{2016}, pp. \bibinfo{pages}{87--102}.
\bibitem[{Van~der Maaten and Hinton(2008)}]{tsne}
\bibinfo{author}{L.~Van~der Maaten}, \bibinfo{author}{G.~Hinton},
\newblock \bibinfo{title}{Visualizing data using t-sne.},
\newblock \bibinfo{journal}{JMLR} \bibinfo{volume}{9} (\bibinfo{year}{2008}).

\end{thebibliography}
